\documentclass{article}

\usepackage{microtype}
\usepackage{graphicx}
\usepackage{subfigure}
\usepackage{booktabs} % for professional tables

\usepackage{hyperref}

\usepackage[accepted]{icml2025}

\usepackage{amsmath}
\usepackage{amssymb}
\usepackage{mathtools}
\usepackage{amsthm}

\usepackage{bm}

\usepackage{tikz}
\usetikzlibrary{matrix,calc}

\usepackage{amsmath}
\usepackage{framed,multirow}
\usepackage{amssymb}
\usepackage{latexsym}
\usepackage{url}
\usepackage{xcolor}
\usepackage{lipsum}

\usepackage[capitalize,noabbrev]{cleveref}

\theoremstyle{plain}
\newtheorem{theorem}{Theorem}[section]

\newtheorem{lemma}[theorem]{Lemma}

\theoremstyle{definition}
\newtheorem{definition}[theorem]{Definition}

\theoremstyle{remark}

\usepackage[textsize=tiny]{todonotes}

\icmltitlerunning{Efficient Noise Calculation in Deep Learning-based MRI Reconstructions}

\begin{document}

\twocolumn[
\icmltitle{Efficient Noise Calculation in Deep Learning-based MRI Reconstructions}



\icmlsetsymbol{equal}{*}

\vskip -0.3in

\begin{icmlauthorlist}
\icmlauthor{Onat Dalmaz}{ee,rad}
\icmlauthor{Arjun D. Desai}{ee,rad}
\icmlauthor{Reinhard Heckel}{tum}
\icmlauthor{Tolga Çukur}{bil}
\icmlauthor{Akshay S. Chaudhari}{bds,rad}
\icmlauthor{Brian A. Hargreaves}{ee,rad,bioe}
\end{icmlauthorlist}

\icmlaffiliation{ee}{Department of Electrical Engineering, Stanford University, Stanford, CA, United States}
\icmlaffiliation{rad}{Department of Radiology, Stanford University, Stanford, CA, United States}
\icmlaffiliation{bds}{Department of Biomedical Data Science, Stanford University, Stanford, CA, United States}
\icmlaffiliation{bioe}{Department of Bioengineering, Stanford University, Stanford, CA, United States}
\icmlaffiliation{tum}{Department of Computer Engineering, Technical University of Munich, Munich, Germany}
\icmlaffiliation{bil}{Department of Electrical and Electronics Engineering, Bilkent University, Ankara, Turkey}

\icmlcorrespondingauthor{Onat Dalmaz}{onat@stanford.edu}

\icmlkeywords{MRI reconstruction, deep learning, noise, uncertainty quantification, Jacobian, sketching, Monte-Carlo, robustness}

\vskip 0.2in
]
\printAffiliationsAndNotice{}

\begin{abstract}
Accelerated MRI reconstruction involves solving an ill-posed inverse problem where noise in acquired data propagates to the reconstructed images. Noise analyses are central to MRI reconstruction for providing an explicit measure of solution fidelity and for guiding the design and deployment of novel reconstruction methods. 
However, deep learning (DL)-based reconstruction methods have often overlooked noise propagation due to inherent analytical and computational challenges, despite its critical importance.
This work proposes a theoretically grounded, memory-efficient technique to calculate \emph{voxel-wise variance} for quantifying uncertainty due to acquisition noise in accelerated MRI reconstructions. 
Our approach approximates noise covariance using the DL network's Jacobian, which is intractable to calculate. To circumvent this, we derive an \textit{unbiased estimator} for the diagonal of this covariance matrix—voxel-wise variance—, and introduce a Jacobian sketching technique to efficiently implement it. We evaluate our method on knee and brain MRI datasets for both data- and physics-driven networks trained in supervised and unsupervised manners. Compared to empirical references obtained via Monte-Carlo simulations, our technique achieves near-equivalent performance while reducing computational and memory demands by an order of magnitude or more.
 Furthermore, our method is robust across varying input noise levels, acceleration factors, and diverse undersampling schemes, highlighting its broad applicability. Our work \emph{reintroduces} accurate and efficient noise analysis as a central tenet of reconstruction algorithms, holding promise to reshape how we evaluate and deploy DL-based MRI. Our code will be made publicly available upon acceptance.
\end{abstract}
\section{Introduction}

%%%%%%%%%%%%%%%%%%%%

Magnetic Resonance Imaging (MRI) has become indispensable in clinical diagnostics, yet prolonged acquisition times can reduce patient throughput and exacerbate motion artifacts. Parallel imaging (pMRI) \cite{sense, grappa} was developed to mitigate these challenges by undersampling k-space and exploiting multiple receiver coils with distinct sensitivity profiles. Despite successfully reducing scan duration, pMRI inevitably introduces spatially varying noise amplification—commonly quantified by the \emph{g-factor}—due to the need to invert an ill-conditioned system. In regions where coil sensitivities overlap or are highly attenuated, the reconstruction leads to locally elevated noise levels \cite{sense, smash_noise}. Early pMRI research rigorously analyzed this spatial noise amplification, culminating in precise formulations that ensured reliable signal-to-noise ratio (SNR) performance to guide algorithm design and clinical implementation \cite{sense, smash_noise, noise_derivation_parallel_imaging}. For instance, the widely used SENSE algorithm explicitly incorporates coil sensitivities to minimize noise, reflecting how classical pMRI approaches typically maintain a clear link between measurement noise and the reconstructed image. 

In contrast, most deep learning (DL) reconstruction methods \cite{raki, knoll_2020, zero_ss,tamir_deep_gen_mri,hwang_joint,Wang_2024_CVPR,grappanet} have not provided an explicit account of how noise from undersampled k-space propagates into the final image. A major obstacle lies in the highly \emph{nonlinear} and complex nature of DL reconstructions \cite{deep_review} making it nontrivial to derive analytical frameworks to characterize how undersampling and learned regularization affect localized noise amplification. In response, the core emphasis in the design of DL methods typically sought maximization of summary metrics like peak signal-to-noise ratio (PSNR) or structural similarity index (SSIM) that do not provide an acute assessment on the spatial noise distribution \cite{deep_robust,adamson2023, fastmri_20}. Thus, the interplay between noise propagation and neural network parameters—and its influence on reconstruction robustness—has received comparatively little attention \cite{Darestani2021,dalmaz2024}.

% As a result, the critical question of whether these networks preserve reliable SNR behavior, especially under varying acceleration factors or heterogeneous noise scenarios, remains an open question 

Nevertheless, characterizing noise in reconstructions is pivotal for advancing DL-MRI, as noise directly impacts SNR—a fundamental image quality metric in MRI that strongly influences clinical utility \cite{grappa_noise,SNR_measure}. Although metrics such as PSNR and SSIM remain common in DL, these metrics provide \emph{image-wide summary measures} of performance and do not capture local variations in noise or SNR across the image \cite{mason2020}. Consequently, important diagnostic regions may be obscured by unpredictable noise amplification patterns that that are either invisible or averaged out when using these global metrics \cite{ssfd,fastmri_19}. In contrast, explicit noise quantification by means of spatial variance maps can yield critical insights into how reconstruction algorithms handle and potentially amplify noise, shaping design decisions to improve robustness and reliability \cite{image_recon_SNR,knoll2019}, paving the way for clinically meaningful performance improvements. In modern unsupervised \cite{ssdu,ddm2} or semi-supervised paradigms \cite{yurt2022}, SNR metrics can be especially valuable by allowing benchmarking reconstructions without relying on fully sampled ground-truth data \cite{iqa_mri}. Moreover, understanding noise propagation can inform tailored sampling schemes—enabling the identification of k-space points that contribute disproportionately to image variability \cite{alkan2024,peng2022_learning} and guide development of noise-aware architectures and training strategies \cite{n2r}. Ultimately, noise characterization strengthens image-quality assessments and fosters the design of DL-based MRI algorithms. Consequently, this increases confidence in deployed models that align with clinical requirements  \cite{akshay_deployment}.

\textbf{Our contributions are threefold:} \textbf{(i)} We provide a comprehensive theoretical framework that provides insights into how acquisition noise in \emph{k}-space propagates to image uncertainty in DL-based MRI reconstructions. \emph{Voxel-wise variance} of reconstructed images is linked to row vectors of DL network’s Jacobian, modulated by k-space correlations and the imaging operator. To circumvent calculation of the full Jacobian, an unbiased estimator for the diagonal elements of the noise covariance matrix is introduced;
\textbf{(ii)} We show how the derived estimator can be implemented through a novel Jacobian sketching algorithm that leverages a complex-valued sketching matrix with random-phase columns, modulated by the adjoint operator and the noise covariance. This approach offers significant advantages over traditional Monte-Carlo (MC) simulation-based methods by reducing computational and memory demands by an order of magnitude or more;
\textbf{(iii)} We rigorously evaluate the proposed method on knee and brain MRI datasets using various network architectures, including data-driven and physics-driven models trained in paradigms ranging from supervised to unsupervised. Experimental results demonstrate that the technique matches (MC) simulation-based empirical references in noise calculation with average correlation $99.8\%$ and error $0.8\%$. We further validate effectiveness of our method across different input noise levels, acceleration factors, and undersampling schemes, showing robustness across clinically relevant imaging scenarios.
Collectively, our method aims to reestablish accurate and efficient noise analysis in reconstruction algorithms, enhancing the evaluation and deployment of DL-based MRI.

% In this study, we propose a novel, comprehensive theoretical framework for quantifying the \emph{voxel-wise variance} of DL-based MRI reconstructions, thereby reintroducing the critical importance of noise analysis. We begin with a first-order Taylor series expansion that expresses the covariance of the reconstructed image in terms of the network’s Jacobian, linking acquisition noise in \emph{k}-space to voxel-level uncertainty. Directly forming the high-dimensional Jacobian is impractical; hence, we introduce a theoretically guaranteed estimator for its diagonal elements (i.e., the per-voxel variances). We further show that this estimator can be implemented efficiently via \emph{sketching the Jacobian}, using a complex-valued sketching matrix with random-phase probing vectors modulated by the adjoint operator and the noise covariance. This leads to an efficient algorithm that can be integrated directly into current automatic differentiation toolsets via Jacobian-vector products. Consequently, our framework both offers insights into noise propagation while remaining scalable and practical for real-world clinical or research workflows in deep reconstructions.

\section{Related Work}
\textbf{Jacobian Sketching in Machine Learning.}
Various works have leveraged Jacobian sketching in machine learning for distinct objectives. For instance, \cite{pmlr-v151-heckel22a,j_sketch_lifelong} apply Jacobian approximations to mitigate catastrophic forgetting in continual learning, while \cite{variance_j_sketch} employs sketching to reduce the variance of stochastic quasi-gradient methods. 
In these settings, the main goal is to approximate or maintain a Jacobian- or Hessian-related quantity to improve optimization or regularize model updates. 
We instead adopt a sketching approach purely for \emph{noise analysis} in MRI—i.e., to estimate the diagonal of the image covariance matrix-rather than for broader optimization or continual learning objectives.

\textbf{Uncertainty Quantification in MRI.}
Prior works on uncertainty quantification (UQ) in MRI generally fall into three categories: 
(1)~classical compressed sensing (using Monte Carlo (MC) reconstructions for confidence intervals) \cite{uq_2021,uq_2024}, 
(2)~Bayesian \cite{bayesian2021} or generative \cite{uq_2021} approaches (e.g., VAEs or i.i.d. Gaussian priors) , and 
(3)~ensemble-based epistemic analyses (training multiple networks in parallel) \cite{kustner_2024}. 
In contrast, our method focuses on \emph{direct noise propagation} from undersampled \emph{k}-space through modern deep architectures, requiring neither repeated sampling nor restrictive assumptions. 
By introducing a Jacobian-based sketching technique, we provide memory- and computation-efficient voxel-wise noise variance estimates for correlated multi-coil settings, targeting the \emph{aleatoric} uncertainty due to acquisition noise.

\textbf{Noise Calculation in DL-MRI.}
Related work on quantifying noise in DL-based MRI reconstructions often employs MC simulations \cite{pmr, simulation, akcakaya2014, dalmaz2024}, repeatedly injecting synthetic perturbations into \emph{k}-space and reconstructing numerous noisy realizations. This approach can produce accurate estimates via a large number of trials which require considerable computational and memory resources---and is thus impractical for large 3D or 4D volumes. Moreover, treating the network as a black box limits interpretability by obscuring how noise propagates or is amplified within the deep reconstruction pipeline. A recent theoretical result \cite{dawood2024} derives analytical noise estimates for \emph{k}-space interpolation networks, and thereby is not generalizable across different architectures.
Furthermore, it involves computing the entire network Jacobian, which quickly becomes impractical for typical 2D or 3D MRI volumes. These limitations highlight the need for a versatile, scalable, and theoretically grounded approach to noise analysis—one that avoids intensive sampling, preserves interpretability, and can be seamlessly integrated into diverse reconstruction paradigms.

\section{Theory}
\label{sec:theory}
% We now provide background on the problem considered as well as neural network based reconstruction approaches. 
\subsection{Accelerated MRI}
\label{subsec:mri_recon_theory}
We acquire \emph{k}-space measurements
\(\bm{y} \in \mathbb{C}^{m}\) from an unknown image \(\bm{x} \in \mathbb{C}^n\)
via a linear \emph{imaging operator} \(\bm{A} \in \mathbb{C}^{m \times n}\), comprising coil sensitivity maps, Fourier encoding, and sampling mask. In practice, \(\bm{y}\) is often undersampled, and corrupted by noise
\(\bm{n}\in \mathbb{C}^{m}\). Thus, the forward model is:
\begin{equation}
\label{eq:forward_model_mri}
\bm{y} 
\;=\; 
\bm{A}\,\bm{x} 
\;+\; 
\bm{n}.
\end{equation}
% Here, the imaging operator \(\bm{A}\) comprises coil sensitivity maps, Fourier encoding, and sampling mask.

% \textbf{Noise Model in Multi-Coil MRI.}
% \label{subsec:multi_coil_noise}
Acquisition noise \(\bm{n}\) is modeled
as complex \emph{Additive White Gaussian Noise (AWGN)} with zero mean:

\begin{equation}
\label{eq:input_noise}
    \bm{n} \;\sim\; \mathcal{CN}\bigl(\bm{0}, \bm{\Sigma}_{k}\bigr),
\end{equation}
where \(\bm{\Sigma}_{k} \in \mathbb{C}^{m \times m}\) is the \emph{sample covariance matrix} of the \emph{k}-space data.  At each \emph{k}-space point, noise is correlated across coils but is independent across distinct \emph{k}-space locations. This implies that a \emph{coil covariance matrix} $\widetilde{\bm{\Sigma}}_{k}$  fully captures the noise statistics. Yet, for simplicity in our linear-algebraic treatment and factorizations, we focus on the Hermitian positive semi-definite (HPSD) \emph{sample covariance matrix} \(\bm{\Sigma}_{k}\in\mathbb{C}^{m\times m}\) (see \textbf{Appendix~\ref{appendix:coil_noise_deriv}}). 

\subsection{Neural Network MRI Reconstruction}
\label{sec:NN_MRI_recon}

In practice, deep MRI reconstructions often follow one of two paradigms:
\textbf{(1)~Unrolled Architectures}, where a series of learned regularization blocks and
data-consistency (DC) steps are iterated for \(K\) cycles \cite{unrolled_1,unrolled_2,fabian2022humusnet}, or
\textbf{(2)~Purely Data-Driven Mappings}, where a feed-forward network performs a direct mapping  \cite{mardani2019}). Despite their structural 
differences, both networks can be viewed as a non-linear function $f\!\Bigl(\bm{A}^H\,\bm{y}\Bigr)$,
where \(\bm{A}^H\) is the adjoint operator that yields a naive zero-filled (ZF) estimate. Without loss of generality, any least-squares or pseudo-inverse initialization could be used (e.g. $\bm{A}^{\dagger}\,\bm{y}$); such choices would simply alter the linear operator involved in subsequent derivations. Thus, our analyses are agnostic to these architectural details, and simply require that \(f\) be differentiable so that its Jacobian exists (See Appendix \ref{appendix:jacobian_existence}).

\subsection{Noise Propagation}
\label{subsec:noise_propagation}
In MRI, we seek to quantify the \emph{voxel-wise variance} of the reconstructed image
\(\bm{x}\in\mathbb{C}^n\) induced by the acquisition noise in \emph{k}-space. Let
\(f:\mathbb{C}^n \to \mathbb{C}^n\) denote a learned \emph{reconstruction function} (e.g., an unrolled network), with Jacobian \(\bm{J}_f(\bm{x}) =\left[\begin{array}{lll}
\frac{\partial \mathbf{f}}{\partial x_1} & \cdots & \frac{\partial \mathbf{f}}{\partial x_n}
\end{array}\right] \in \mathbb{C}^{n\times n}\).

DL pipelines commonly initialize \(\bm{x}^{(0)} = \bm{A}^{H}\,\bm{y}\) with the "noisy" measured data \(\bm{y} = \bm{y}_0 + \bm{n}\), where \(\bm{y}_0 = \bm{A}\,\bm{x}\). Then:
\begin{align}
\label{eq:initial_estimate}
    \bm{x}^{(0)} 
=
\bm{A}^{H}\,\bigl(\bm{y}_0 + \bm{n}\bigr)
=
\bm{A}^{H}\,\bm{y}_0 
+
\bm{A}^{H}\,\bm{n}
=
\bm{x}_0
+
\bm{A}^{H}\,\bm{n},
\end{align}
with \(\bm{x}_0 = \bm{A}^{H}\,\bm{y}_0\) indicating the noise-free component of this initial
estimate. A \emph{first-order Taylor expansion} of the reconstruction function
\(f(\cdot)\colon\mathbb{C}^n \to \mathbb{C}^n\) around \(\bm{x}_0\) approximates:
\begin{equation}
\label{eq:taylor}
f\bigl(\bm{x}^{(0)}\bigr) 
\;\approx\; 
f(\bm{x}_0) 
\;+\; 
\bm{J}_f(\bm{x}_0)\,\bigl(\bm{x}^{(0)} - \bm{x}_0\bigr),
\end{equation}
where \(\bm{J}_f(\bm{x}_0)\in \mathbb{C}^{n\times n}\) is the Jacobian matrix of \(f\) evaluated
at \(\bm{x}_0\). From \ref{eq:initial_estimate} and \ref{eq:taylor} the perturbation in the reconstructed image due to the noise
\(\bm{n}\) is:
\[
\delta \bm{x} 
=
f\bigl(\bm{x}^{(0)}\bigr) - f(\bm{x}_0)
\;\approx\;
\bm{J}_f(\bm{x}_0)\,\bigl(\bm{A}^H \bm{n} \bigr)
\]
Note that \(\delta \bm{x}\) also remains zero-mean:

\begin{equation}
    \mathbb{E}\bigl[ \delta \bm{x}\bigr] 
\;=\;
\bm{J}_f(\bm{x}_0)\,\bm{A}^H \mathbb{E}\bigl[  \bm{n}\bigr] \;= \bm0.
\end{equation}
It is also worthwhile to note that due to linearity, approximate reconstruction noise shares the same Gaussian nature, with the covariance matrix:
\begin{align}
\label{eq:covariance_image}
\bm{\Sigma}_{\bm{x}}
&=\;
\mathbb{E}\bigl[\delta \bm{x}\,\delta\bm{x}^H\bigr]
=\;
\bm{J}_f(\bm{x}_0)\,\bm{A}^{H}
\,\mathbb{E}[\bm{n}\,\bm{n}^H]
\,\bm{A}\,\bm{J}_f(\bm{x}_0)^{H} \nonumber \\
&=\;
\bm{J}_f(\bm{x}_0)\,\bm{A}^{H}
\,\bm{\Sigma}_k
\,\bm{A}\,\bm{J}_f(\bm{x}_0)^{H}.
\end{align}

Even though the Jacobian \(\bm{J}_f=\bm{J}_f(\bm{x}_0)\in\mathbb{C}^{n\times n}\) exists, storing or explicitly computing the full matrix is
intractable for MR images, due to dimensionality and massive data size. Nonetheless,  \eqref{eq:covariance_image} shows propagation of k-space
covariance \(\bm{\Sigma}_k\) through the adjoint operator \(\bm{A}^{H}\) and the
network’s Jacobian \(\bm{J}_f\). Similar constructs for the noise matrix exist in other pMRI
approaches \cite{sense}.

\textbf{Factorization and Diagonal Entries.}
To simplify computations, we first perform Cholesky decomposition of $\bm{\Sigma}_k$:
\begin{align}
    \bm{\Sigma}_k = \bm{\sigma}_k \,\bm{\sigma}_k^H,
\end{align}
where $\bm{\sigma}_k$ is the Cholesky factor of $\bm{\Sigma}_k$.
Substituting into \eqref{eq:sigma_x_definition}, we obtain
\begin{align}
\label{eq:sigma_x_definition}
\bm{\Sigma}_{\bm{x}}
&= \bm{J}_f \bm{A}^H \bigl( \bm{\sigma}_k \bm{\sigma}_k^H \bigr) \bm{A} \bm{J}_f^{H}, \\
&= \Bigl(\bm{J}_f\,\bm{A}^H\,\bm{\sigma}_k\Bigr)
\Bigl(\bm{J}_f\,\bm{A}^H\,\bm{\sigma}_k\Bigr)^H.
\end{align}
Here, we see that \(\bm{\Sigma}_{\bm{x}} = \bm{L}\,\bm{L}^H\), with Cholesky factor:
\begin{align}
\label{eq:sigma_x_cholesky}
\bm{L}
~\;=\;~
\bm{J}_f\,\bm{A}^H\,\bm{\sigma}_k,
\end{align}

The variance of the $i$-th voxel (\textit{i.e.}, the $i$-th diagonal entry of $\bm{\Sigma}_{\bm{x}}$) is then:
\begin{align}
\label{eq:naive_calculation}
\mathrm{Var}(\bm{x}_i) 
\;=\; 
 \mathrm{diag}(\bm{\Sigma}_{\bm{x}})_i
\;=\;
[\bm{L}\,\bm{L}^H]_{ii}
\;=\;
\|\bm{l}_i\|_2^2,
\end{align}
%naive_calculation
where $\bm{l}_i \in \mathbb{C}^n$ is the $i$-th \textit{row} of $\bm{L}$:
\[
\bm{l}_i
=
\nabla^{\top} f_i\,\bm{A}^H\,\bm{\sigma}_k
\]
In principle, the variance of each voxel could be iteratively obtained via taking the $\ell_2$ norm of $i$-th row of Jacobian, transformed and scaled by \(\bm{A}^H\) and \(\bm{\sigma}_k\).  Although this direct approach would be computationally heavy (see \textbf{Algorithm} \ref{algo:naive_calc}~in Appendix), the analysis reveals how \(\bm{\Sigma}_k\), $\bm{A}$ and \(\bm{J}_f\) together determine the voxel-wise noise distribution.

\subsection{Estimating Noise via Jacobian Sketching}
\label{sec:variance_estimation}

Here we present an algorithm for estimating the diagonal of the covariance matrix
$\bm{\Sigma}_{\bm{x}}$ in an unrolled MRI reconstruction, \emph{without} explicitly forming or
storing the Jacobian $\bm{J}_f \in \mathbb{C}^{n\times n}$ or the image covariance
$\bm{\Sigma}_{\bm{x}} \in \mathbb{C}^{n\times n}$. We instead rely on
\emph{Jacobian-Vector Products (JVPs)}, enabling a \emph{sketch-based} approach that efficiently probes the
Jacobian and covariance structure through random vectors.
We first present the following results regarding the diagonal entries of implicit, complex-valued Hermitian covariance matrices.

\begin{theorem}[\textsc{Unbiased Diagonal Estimator}]
\label{thm:unbiased_estimator}
Let \(\bm{\Sigma}\in \mathbb{C}^{n\times n}\) be Hermitian, and let \(\bm{v}\in\mathbb{C}^n\) 
satisfy
\begin{equation}
\label{eq:conditions_complex}
\mathbb{E}[\bm{v}] \;=\; \bm{0},
\quad
\mathbb{E}[\bm{v}\,\bm{v}^{H}] \;=\; \bm{I}.
\end{equation}
Define \(\bm{y} = (\bm{\Sigma}\,\bm{v}) \odot \bm{v}^{*}\), where $\odot$  is the Hadamard product, and $*$ is scalar complex conjugation. Then:

\[\mathbb{E}[\bm{y}] = \mathrm{diag}\bigl(\bm{\Sigma}\bigr)\]
\end{theorem}

\begin{proof}
\textbf{
$
\bm{a} \odot\bm{b}
=
\mathrm{diag}\bigl( \bm{a} \bm{b^T}\bigr).
$
}

Write \(\bm{\Sigma}=[\Sigma_{ij}]\) and \(\bm{v}=(v_1,\dots,v_n)^\top\). Then for each index $i$:
\[
 (\bm{\Sigma}\,\bm{v})_i
\;=\;
\sum_{j=1}^n \Sigma_{ij}\,v_j.
\]
Multiplying by \(v_i^{*}\) yields $y_i$:
\[
y_i = (\bm{\Sigma}\,\bm{v})_i\,v_i^{*}
\;=\;
\sum_{j=1}^n
\Sigma_{ij}\,v_j\,v_i^{*}.
\]
Taking expectation:
\[
\mathbb{E}[y_i]
\;=\;
\mathbb{E}\Bigl[\sum_{j=1}^n
\Sigma_{ij}\,v_j\,v_i\bigr)^{*}] \;=\;
\sum_{j=1}^n
\Sigma_{ij}\,\mathbb{E}\Bigl[v_j\,v_i^{*}\Bigr]
\]

From \ref{eq:conditions_complex}, we note  \(\mathbb{E}[\,v_j\,v_i^{*}\,] = \delta_{ij}\):
\[
\mathbb{E}[y_i] = \sum_{j=1}^n
\Sigma_{ij}\,\delta_{ij} \;=\;
\Sigma_{ii}.
\]
Since this holds for all \(i\), we have 
\(\mathbb{E}[\bm{y} ] = \mathrm{diag}(\bm{\Sigma})\).
\end{proof}
\begin{lemma}
\label{lem:variance_estimator}
Let $\bm{\Sigma}\in\mathbb{C}^{n\times n}$ be HPSD with a Cholesky factorization $\bm{\Sigma} = \bm{L}\,\bm{L}^{H}$.
Suppose $\bm{v}\in\mathbb{C}^{n}$ satisfies \eqref{eq:conditions_complex}. Defining $\bm{u}=\bm{L}\,\bm{v}$, we claim:
\[
\mathbb{E}\bigl[\bm{u}\odot\bm{u}^*\bigr]
=
\mathrm{diag}\bigl(\bm{\Sigma}\bigr).
\]
\end{lemma}

\begin{proof}
Since $\bm{u} = \bm{L}\,\bm{v}$, its $i$th component is
\[
u_i = \sum_{j=1}^n L_{ij}\,v_j.
\]
Hence, the $i$th entry of the Hadamard product $\bm{u}\odot\bm{u}^*$ is
\[
[\bm{u}\odot\bm{u}^*]_i
\,=\,
u_i\,(u_i)^*
\,=\,
\Bigl(\sum_{j=1}^n L_{ij}\,v_j\Bigr)
\Bigl(\sum_{k=1}^n L_{ik}^{*}\,v_k^{*}\Bigr).
\]
Rewriting and taking expectation:
\[
\mathbb{E}\bigl[u_i\,(u_i)^*\bigr]
=
\sum_{j=1}^n
\sum_{k=1}^n
L_{ij}\,L_{ik}^{*}\,
\mathbb{E}[v_j\,v_k^{*}].
\]
By \eqref{eq:conditions_complex}, \(\mathbb{E}[v_j\,v_k^{*}] = \delta_{jk}\). Thus,
\[
\mathbb{E}\bigl[u_i\,(u_i)^*\bigr]
=
\sum_{j=1}^n
L_{ij}\,L_{ij}^{*}
=
[\bm{L}\,\bm{L}^H]_{ii}
=
\Sigma_{ii}.
\]
Because this holds for each $i$, we conclude
\[
\mathbb{E}[\bm{u}\odot\bm{u}^*]
=
\mathrm{diag}(\bm{\Sigma}).
\]
\end{proof}
Returning to \eqref{eq:sigma_x_cholesky}, we have, $\bm{L} =\bm{J}_f\,\bm{A}^{H}\,\bm{\sigma}_k$. Then, by Lemma~\ref{lem:variance_estimator}, 
\[
\bm{u}
=
\bm{L}\,\bm{v}
=
\bm{J}_f \Bigl(\bm{A}^{H}(\bm{\sigma}_k\,\bm{v})\Bigr) \Rightarrow
\mathbb{E}\bigl[\bm{u} \odot \bm{u}^* \bigr]
=
 \mathrm{diag}(\bm{\Sigma}_{\bm{x}}).
\]
Thus, applying $\bm{J}_f$ to suitably distributed random vectors reveals each diagonal entry of
\(\bm{\Sigma}_{\bm{x}}\). 

\subsubsection{Vectorized Implementation: Jacobian Sketching}
\label{sec:jacobian_sketch_vectorized}
A naive approach would sample each $\bm{v}_j\in\mathbb{C}^m$, compute 
$\bm{L}\,\bm{v}_j=\bm{J}_f(\bm{A}^H\bm{\sigma}_k\,\bm{v}_j)$ individually, and 
accumulate $\bm{u}_j$. Instead, we provide a \emph{vectorized} algorithm that improves computational and efficiency (See Algorithm~\ref{alg:variance_estimation_matrix}):

\begin{enumerate}
    \item \textbf{Generate random matrix:} 
          $\bm{V}_S \in \mathbb{C}^{m\times S}$ with columns $\bm{v}_j$  s.t. $\mathbb{E}[\bm{v}_j] = \bm{0}$, and
          $\mathbb{E}[\bm{v}_j\,\bm{v}_j^H]=\bm{I}_m$.
    \item \textbf{Transform by $\bm{\sigma}_k$ and $\bm{A}^H$:} 
          $\bm{W}_S = \bm{\sigma}_k\,\bm{V}_S$, then 
          $\widetilde{\bm{W}}_S = \bm{A}^H\,\bm{W}_S$
    \item \textbf{Sketch $\bm{J}_f$ via $\widetilde{\bm{W}}_S$} :
          $\bm{U}_S = \bm{J}_f \widetilde{\bm{W}}_S \in\mathbb{C}^{n\times S}$.
    \item \textbf{Hadamard product w/Hermitian \& Average:} 
          $\bm{V}_{\text{samples}}=\bm{U}_S \odot \bm{U}^H_S\in\mathbb{R}^{n\times S}$,  \\
          $\widehat{\mathrm{diag}(\bm{\Sigma}_{\bm{x}})} = \frac{1}{S}\,\bm{V}_{\text{samples}}\,\mathbf{1}_S$.
\end{enumerate}

\subsubsection{Choice of Random Vectors}
\label{subsec:choice_random_vectors}
For unbiased estimation, we require random vectors \(\bm{v}\in\mathbb{C}^n\) satisfying  (\ref{eq:conditions_complex}). A natural design choice would be \emph{standard complex Gaussian} vectors where each $v_i$ is drawn from $\sim \mathcal{CN}(0,1)$ independently. Yet, in \emph{real}-valued matrix diagonal estimation problems, 
\emph{Rademacher} vectors (i.e., \(\pm1\)) have been shown to reduce estimator variance compared to Gaussian vectors \cite{hutchinson,bekas}. 
In MR reconstruction, however, data and operators are inherently \emph{complex}, thus we propose using a \emph{complex} analogue, namely
\emph{complex Rademacher (random-phase)} vectors, where each \(v_i\) has unit magnitude and uniformly random phase:
\[v_i = e^{j\,\theta_i},\;\theta_i \sim \mathrm{Uniform}[0,2\pi].\]
We show in Appendix \ref{appendix:complex_variance} that both choices satisfy (\ref{eq:conditions_complex}),  though \emph{random-phase} vectors yield strictly lower \emph{estimator variance}; ergo selected for consequent experiments.

\section{Experiments and Results}
\label{sec:experiments}
\subsection{Datasets and Experimental Setup}
\label{subsec:datasets_experimental}
We performed experiments on two publicly available datasets—\href{https://old.mridata.org/fullysampled/knees}{\textbf{Stanford knee dataset}}
 and a subset of the \textbf{fastMRI brain dataset} \cite{fastmri}. The Stanford knee dataset consists of 8-channel 3D FSE PD-weighted scans, which we split into 14 subjects for training, 2 for validation, and 3 for testing. Each 3D scan was demodulated and decoded via a 1D inverse Fourier transform along the readout dimension, yielding 2D Axial slices with matrix size $320\times256$.
The fastMRI brain dataset comprises 16-channel Axial T2-weighted scans with matrix size $384\times384$, split into 54 training, 20 validation, and 30 testing subjects. We focus on using these test volumes primarily to evaluate the variance estimation performance of our framework, rather than to compare the reconstruction quality of different methods. k-space data were retrospectively undersampled via 2D Poisson Disc undersampling masks; and coil sensitivities were estimated via J-SENSE \cite{jsense}. We estimated the coil covariance matrix $\widetilde{\bm{\Sigma}}_{k}$ for each slice using the outermost 5\% of \emph{k}-space points, where signals are minimal \cite{pmr,sense}. During inference, a unique and deterministic undersampling mask was used on each test slice for reproducibility.
\begin{figure*}[t!]
\vskip 0.2in
\begin{center}
\centerline{\includegraphics[width=2\columnwidth]{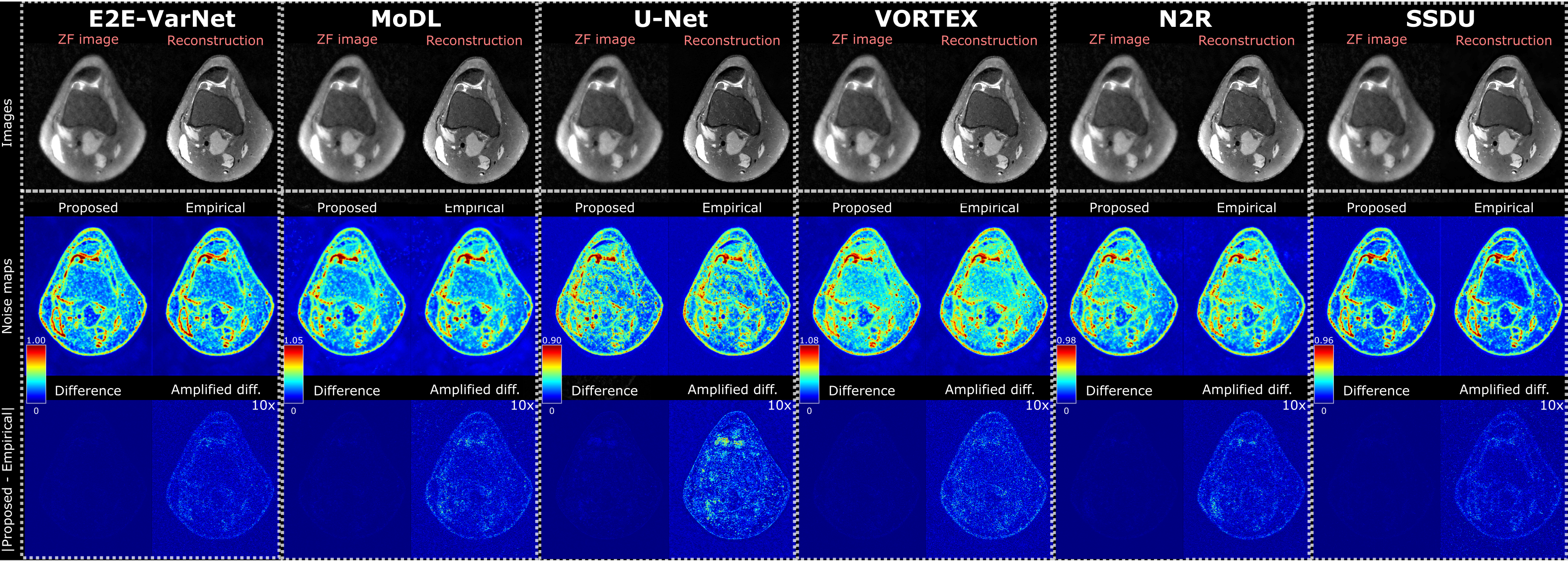}}
\caption{Each column corresponds to a distinct deep reconstruction method at $R=8,\alpha=1$ on knee data. In each column:
    \textbf{(top row)} shows ZF and reconstructed images;
    \textbf{(middle row)} displays noise variance maps derived by the proposed
    method and empirical simulations;
    \textbf{(bottom row)} presents difference and amplified (x10) difference maps between the proposed and empirical variance maps to highlight spatial discrepancies. Color bars indicate each panel's relative noise map display window.}
\label{fig:knee_methods}
\end{center}
\vskip -0.2in
\end{figure*}

\textbf{Reconstruction Methods}
\label{subsec:recon_methods}
To comprehensively evaluate our approach, we selected six deep MRI reconstruction algorithms 
spanning different learning paradigms: supervised (\textbf{E2E-VarNet}~\cite{e2e_varnet}, \textbf{MoDL}~\cite{modl}, \textbf{U-Net} \cite{unet}), semi-supervised (\textbf{N2R (Noise2Recon)}, \textbf{VORTEX} \cite{vortex}), self-supervised (\textbf{SSDU}~\cite{ssdu}); and network 
architectures: fully data-driven (\textbf{U-Net} \cite{unet}) vs. physics-driven unrolled (the rest). These methods cover a broad spectrum of deep reconstruction models for examining the generalizability of the proposed noise analysis framework across different architectures and learning paradigms.  Additional details regarding methods including training, hyperparameters, architectures, and results are provided in Appendix \ref{appendix:recon_methods}.

\textbf{Simulating Noise Levels.}
To investigate performance under varying noise scenarios, we simulate different input noise levels by scaling the estimated covariance matrix $\bm{\Sigma}_{k} = 
\alpha \widetilde{\bm{\Sigma}}_{k}$, where is $\alpha \in \{1, 5,10, \cdots,200\}$ so that $\bm{n} \sim \mathcal{N}\bigl(\bm{0},\,\bm{\Sigma}_{k}\bigr)$. 

\textbf{Reference Empirical Baseline.}
For each experimental setting (noise level, acceleration rate, dataset) and image, we performed $N=3,000$ MC trials (See Appendix \ref{appendix:monte}), each time adding simulated noise $\bm{n}$ drawn from $\bm{\Sigma}_k$. We reconstruct each noisy measurement and compute \emph{empirical} variance maps by measuring voxelwise sample variance across all trials. The empirical variance maps serve as a gold standard reference.

\textbf{Variance Estimation.}
 Variance maps computed by our proposed \emph{randomized Jacobian sketching} algorithm were compared to the empirical variance maps (our benchmark) to assess the accuracy of the proposed method for capturing voxel-wise noise propagation. The sketching matrix $\bm{V}_S$ had $S=1000$ column probing vectors, which provides a balance between the estimation accuracy and computation (See Appendix \ref{appendix:sketching_matrix}). Percent (\%) Pearson Correlation Coefficient (PCC), 
and Normalized Root-Mean-Square Error (NRMSE) between computed and empirical variance maps were measured, and mean$\pm$std across all test slices were calculated to quantify the performance of our method. Significance of the differences between voxel-wise noise distributions was assessed by using a two-sample t-test.

\subsection{Generalizability Across Reconstruction Methods}
\label{subsec:results}
To assess how well our variance estimator generalizes across different DL-based reconstruction methods, we fix $R=8$ and $\bm{\Sigma}_{k} = \widetilde{\bm{\Sigma}}_{k}$, then train each model (knee and brain) and estimate noise variance. Table~\ref{table:metrics_comparison} shows that our method achieves near-perfect correlations to the gold standard, with no statistically significant difference betweeen noise distributions ($p<0.05$). Although the U-Net yields slightly higher NRMSE (1.7\%), these values remain small in absolute terms. One contributing factor to this might be U-Net's lack of iterative data-consistency, which applies a single-pass nonlinear denoising. This leads to both lower-fidelity reconstructions (See Table \ref{table:recon_metrics_comparison_merged} in Appendix), and deviations from our method's linear approximation, resulting in minor local estimation errors. 

% Nonetheless, spatial correlations remain extremely high ($\mathrm{PCC}\gtrsim 99.4\%$), indicating that our estimator reliably captures the noise distribution across the image.
Figures~\ref{fig:knee_methods} and \ref{fig:brain_methods} (Appendix) illustrate representative noise-variance maps from knee and brain slices, respectively.
Our estimator generally reproduces the intensity ranges and spatial patterns observed in the empirical reference, showing only minor deviations near regions of sharp intensity transitions, such as tissue boundaries or areas with rapid signal variation. These localized discrepancies remain negligible relative to the overall noise distribution and do not affect the overall reliability of the framework. Across a variety of DL models, our estimator consistently provides robust noise estimates that align well with empirical simulations.

\begin{table}[t]
    \centering
    \small
    \caption{
        Mean$\pm$std PCC and NRMSE between noise-variance maps calculated by our method and by reference empirical baseline, across different DL reconstruction methods for knee and brain data at $R=8, \,\alpha=1$. 
    }
    \label{table:metrics_comparison}
    \setlength{\tabcolsep}{2.3pt} % Adjust column spacing as desired
    \renewcommand{\arraystretch}{1.0} % Adjust row spacing as desired
    \begin{tabular}{lcccc}
        \toprule
        \textbf{Method} & \multicolumn{2}{c}{\textbf{Knee}} & \multicolumn{2}{c}{\textbf{Brain}} \\
        \cmidrule(lr){2-3} \cmidrule(lr){4-5}
        & \textbf{PCC (\%)} & \textbf{NRMSE (\%)} & \textbf{PCC (\%)} & \textbf{NRMSE (\%)} \\
        \midrule
        \textbf{E2E-VarNet} & $99.9 \pm 0.0$ & $0.7 \pm 0.0$ & $99.9 \pm 0.0$ & $0.5 \pm 0.1$ \\
        \textbf{MoDL}       & $99.9 \pm 0.0$ & $0.5 \pm 0.0$ & $99.7 \pm 0.0$ & $1.1 \pm 0.1$ \\
        \textbf{U-Net}      & $99.4 \pm 0.0$ & $1.7 \pm 0.2$ & $99.7 \pm 0.0$ & $1.8 \pm 0.2$ \\
        \textbf{VORTEX}     & $99.9 \pm 0.0$ & $0.6 \pm 0.0$ & $99.9 \pm 0.0$ & $0.6 \pm 0.1$ \\
        \textbf{N2R}        & $99.9 \pm 0.0$ & $0.6 \pm 0.0$ & $99.9 \pm 0.0$ & $0.7 \pm 0.1$ \\
        \textbf{SSDU}       & $99.9 \pm 0.0$ & $0.4 \pm 0.0$ & $99.8 \pm 0.0$ & $0.9 \pm 0.1$ \\
        \bottomrule
    \end{tabular}
    \vskip -0.3in
\end{table}

\begin{table}[t]
\small
    \centering
    \caption{Computational and memory efficiency (per slice) on knee dataset for standard unrolled architecture, i.e. as used in E2E-VarNet, VORTEX, N2R, SSDU.}
    \label{tab:efficiency_comparison}
    \begin{tabular}{lccc}
    \toprule
    \textbf{Metric} & \textbf{Empirical} & \textbf{Proposed} & \textbf{Naive} \\
    \midrule
    \textbf{Time} & 54.0 s & 1.3 s & 880.2 s \\
    \textbf{Storage} & 3513.3 MB & 1.1 MB & 1.1 MB \\
    \bottomrule
    \end{tabular}
    \vskip -0.3in
\end{table}

\subsection{Computational and Memory Benefits}
Table~\ref{tab:efficiency_comparison} compares the computational time and memory usage for three 
variance-estimation strategies for a single slice on the unrolled architecture from the knee dataset: \emph{Empirical}, \emph{Proposed}, and \emph{Naive} (based on Eq. \ref{eq:naive_calculation}). \emph{Empirical} simulations take about a minute and require 
storing multiple reconstructions. In contrast, \emph{Naive} avoids large 
storage overhead, but its exhaustive row-by-row Jacobian calculation is significantly more 
time-consuming. In comparison, our \emph{proposed} estimator balances the two ends: reducing the computation by over an order of magnitude relative to the empirical baseline by leveraging JVPs, while circumventing storage of reconstructed images. These results illustrate how the proposed algorithm attains substantial speed and memory advantages, making it a practical tool for routine use, especially on large datasets. See Appendix \ref{appendix:computation_vs_complexity} for efficiency/scalability of our method with regards to network complexity and architecture, and Appendix \ref{appendix:mc_vs_sketching} for a practical comparison with the empirical baseline at low trial counts.
% further demonstrating the advantages of our approach in both accuracy and speed.

\subsection{Generalization to Various Undersampling Schemes}
\label{subsec:undersampling_effect}
For the baseline noise level ($\alpha=1$), we vary the acceleration factor 
$R\in\{4,8,12,16,24\}$, training a separate model for each $R$. 
Figures \ref{fig:knee_R_variance_maps}, \ref{fig:brain_R_variance_maps} (Appendix) illustrate variance maps for representative knee and brain slices, respectively. Although higher acceleration factors $R$ decrease the number of 
\emph{k}-space samples—thus elevating ill-posedness and noise—our noise estimations continue to align closely with empirical references across all $R$, remaining their accuracy over a broad range of undersampling rates.

% This indicates that while higher acceleration leads to increased image-domain noise, 
% the proposed estimator remains 

\begin{figure}[t!]
% \vskip 0.2in
\begin{center}
\centerline{\includegraphics[width=\columnwidth]{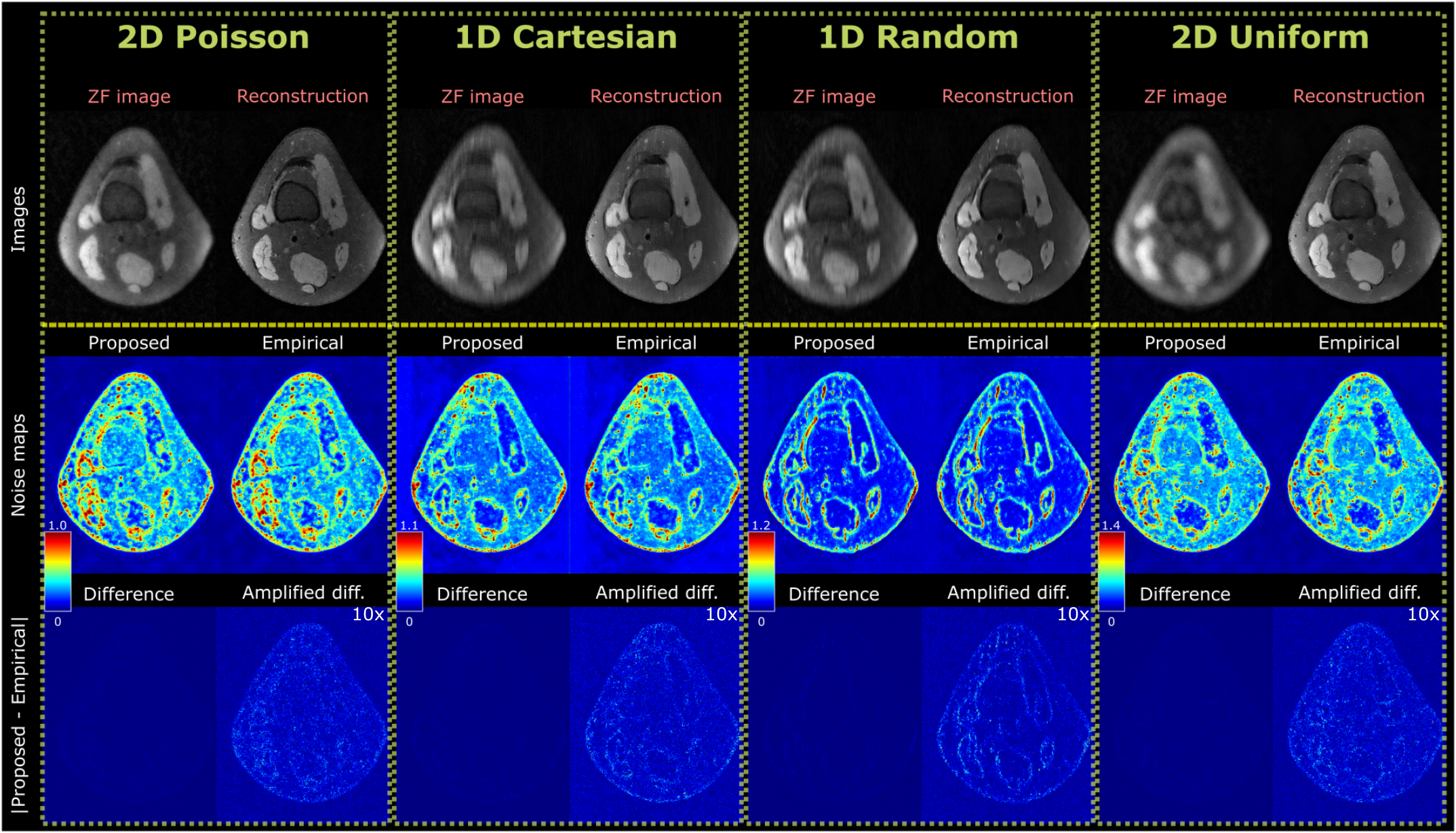}}
\caption{Each column corresponds to a different undersampling pattern (E2E-VarNet on knee data at $R=8, \alpha=1$). In each column:
    \textbf{(top row)} shows ZF and reconstructed images;
    \textbf{(middle row)} displays noise variance maps derived by the proposed
    method and empirical simulations;
    \textbf{(bottom row)} presents difference and amplified (x10) difference maps between the proposed and empirical variance maps to highlight spatial discrepancies. Color bars indicate each panel's relative noise map display window.}
\label{fig:sampling_masks_comparison}
\end{center}
\vskip -0.5in
\end{figure}

We also assessed the generalizability of our method under varying undersampling patterns, 
including 1D uniform Cartesian and 1D random Cartesian undersampling, and 2D uniformly random and Poisson-disc (default sampling pattern used in this study) undersampling 
(see Figure \ref{fig:undersampling_masks} in Appendix). For each pattern, we train a dedicated model, 
then estimate variance in representative slices. Figure~\ref{fig:sampling_masks_comparison} 
demonstrates that our noise maps retain strong agreement with empirical results under 
all sampling schemes—whether uniform or random. This consistency arises because 
the operator $\bm{A}$ inherently encodes the sampling mask; 
our method only relies on $\bm{A}$ as an implicit linear operator, 
making it \emph{undersampling-agnostic}. These results demonstrate robustness to substantial changes in sampling pattern and acceleration rate, underscoring its versatility in diverse imaging scenarios.

\begin{figure*}[ht]
% \vskip 0.2in
\begin{center}
\centerline{\includegraphics[width=2\columnwidth]{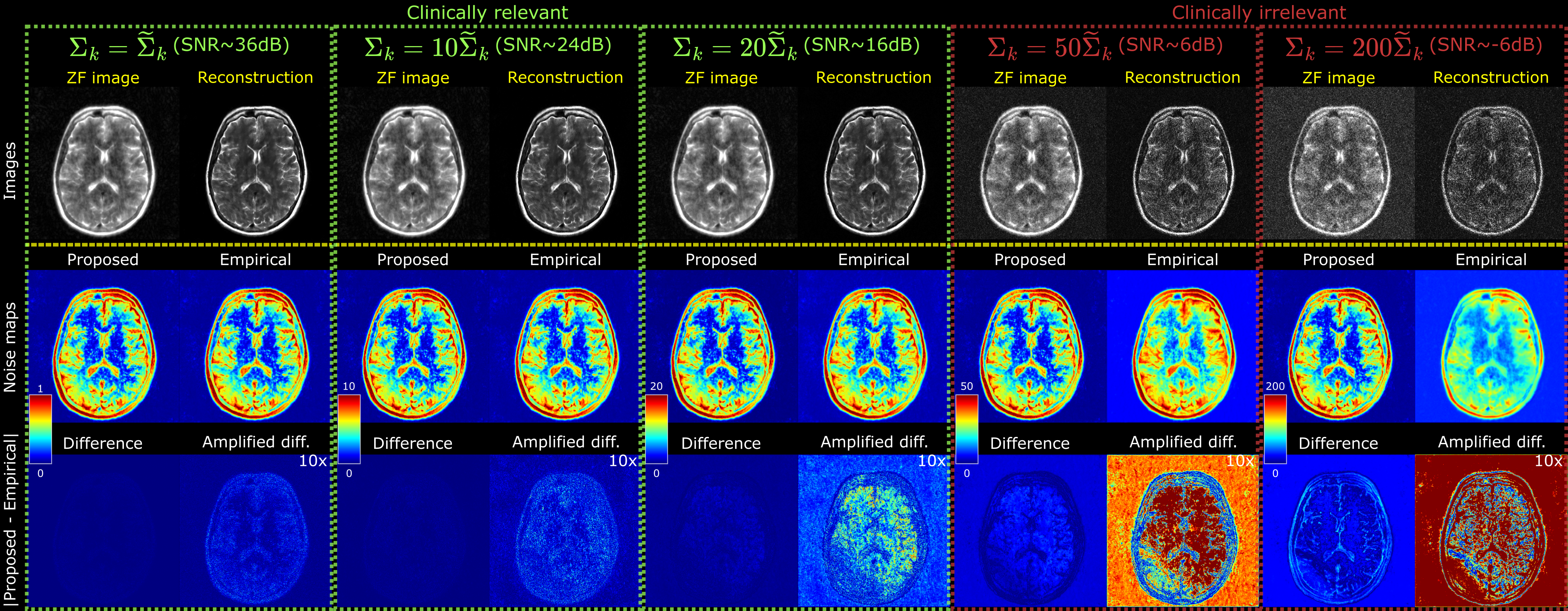}}
\caption{Each column corresponds to a different noise scaling factor $\alpha$ resulting in varying SNR scenarios (E2E-VarNet on knee data at $R=8
$). In each column:
    \textbf{(top row)} shows ZF and reconstructed images;
    \textbf{(middle row)} displays noise variance maps derived by the proposed
    method and empirical simulations;
    \textbf{(bottom row)} presents difference and amplified (x10) difference maps between the proposed and empirical variance maps to highlight spatial discrepancies. Color bars indicate each panel's relative noise map display window. Note that $\alpha=50,200$ corresponds to SNR values well under a 10dBs, which is often cited as a threshold for diagnostic utility \cite{brown2014magnetic}.}
\label{fig:brain_noise_variance_maps}
\end{center}
\vskip -0.4in
\end{figure*}

\subsection{Robustness to k-space Noise Level}
We systematically increase the k-space noise magnitude by varying $\alpha$ as described in \ref{subsec:datasets_experimental}, thereby simulating growing covariance matrices (and thus SNR levels). Figures~\ref{fig:brain_noise_variance_maps}, \ref{fig:noise_analysis_knee} (Appendix) show variance maps for representative brain and knee slices from the proposed method and from empirical simulations, and Figure~\ref{fig:nrmse_vs_noise} (Appendix) shows the error between them as alpha grows. Although the reconstructed images become visibly noisier at higher $\alpha$, networks still perform noticeable denoising. 
As $\alpha$ increases, empirical maps spread variability more uniformly across tissues, leading to a more homogeneous noise pattern without accentuating different tissue types. For moderate noise scaling (\(\alpha < 50\)), the proposed and empirical variance maps closely match. Note that extreme noise levels ($50\times$–$200\times$) correspond to SNR values well below 10–15 dB and thus far beyond what is considered clinically relevant in routine practice \cite{brown2014magnetic,WestbrookMRIinPractice}. At these regimes, differences emerge in low-intensity tissue or around sharp edges, reflecting how the network’s learned regularization systematically suppresses part of the injected noise compared to our estimation. Nevertheless, these difference maps remain small under practical SNR conditions, indicating that our approach effectively captures noise propagation 
for typical or moderately elevated noise.
We provide an additional analysis and discussion in Appendix \ref{appendix:noise_discussion}.
% Figure~\ref{fig:nrmse_vs_noise} further quantifies these observations. At near-baseline noise (\(\alpha\approx 1\)), 
% the error is minimal, indicating accurate variance estimation under real-world scenarios. 
% As \(\alpha\) climbs by one or two orders of magnitude, the error gradually increases, reflecting 
% that the first-order Taylor approximation and linear assumptions can become less valid at 
% severe noise intensities. Despite this growth, the error remains small within the range of 
% noise factors most relevant to clinical imaging, underscoring the robustness and practical 
% usefulness of our variance estimator for moderate noise scales.

% This discrepancy can be attributed to the \emph{denoising} behavior of deep 
% reconstruction networks, which naturally suppress part of the injected noise at higher SNR 
% degradations. Consequently, the empirical STD is systematically lower than a pure first‐order 
% variance propagation might predict. 

\section{Discussion and Conclusion}
% \paragraph{Computational and Memory Efficiency.}
% Because each JVP $\bm{J}_f(\widetilde{\bm{w}}_j)$ avoids forming $\bm{J}_f\in\mathbb{C}^{n\times n}$, 
% we only spend $\mathcal{O}(S\cdot \mathrm{JVP\_cost})$ plus overhead for matrix‐vector multiplications 
% ($\bm{\sigma}_k$, $\bm{A}$, $\bm{A}^H$).  Memory usage is similarly reduced, since we never store 
% $\bm{J}_f$ or $\bm{\Sigma}_k$ in full.  In many practical MRI settings, $S$ is small (tens to hundreds),
% and this approach scales well to large $n$ ($10^5$--$10^6$ voxels).  We can process partial 
% samples in minibatches if needed, further limiting memory demands.  Consequently, 
% \emph{randomized Jacobian sketching} provides a feasible route to estimate voxelwise noise 
% variances for unrolled MRI reconstructions without incurring $\mathcal{O}(n^2)$ complexity or 
% storage.

Our localization at \(\bm{x}_0\) presupposes that \(\|\bm{A}^H \bm{n}\| \ll \|\bm{x}_0\|\)—i.e., the noise perturbation in the image domain must not significantly deviate from the baseline signal, for the first‐order Taylor approximation to remain valid. Clinically, typical MRI acquisitions usually have sufficiently high SNR for this condition to hold, so the linear approximation remains valid \cite{brown2014magnetic,WestbrookMRIinPractice}. However, once noise levels become so large that \(\|\bm{A}^H \bm{n}\|\) approaches or exceeds \(\|\bm{x}_0\|\), the nonlinearities in \(f\) can no longer be ignored. In such extreme cases—such as noise magnitudes exceeding the baseline signal—our method may under‐ or over‐estimate variance, as the learned regularization could reshape these large perturbations in ways that can not be accounted for by a linear approximation. Despite these nuances at high noise regimes, the strong agreement between our method and empirical simulations under realistic acquisition conditions highlights the practical value and efficacy of our approach.

Noise distribution in fully-sampled or linear reconstructions tends to be fairly uniform, modulated by coil sensitivities \cite{pmr}. However, different DL reconstructions exhibit distinct noise profile for the same slice, shaped by factors such as regularization, optimization criteria, and the network's learned biases. Moreover noise distribution often depends on the underlying anatomy and k-space data, rather than being uniformly distributed. Indeed, inspection of the noise maps suggests that spatial noise amplification in DL reconstructions mirrors features in ZF inputs and final outputs. This localized noise amplification has important clinical implications. In routine MRI assessments, radiologists focus on specific regions of interest (ROI) such as cartilage in knee scans or pathology-affected areas in the brain, rather than examining the entire field of view uniformly. Therefore, even if global metrics indicate high reconstruction quality, a locally elevated noise variance in diagnostically relevant ROIs could compromise clinical interpretation. The ability to accurately quantify noise and preserve reliable diagnostic signal in crucial ROIs is pivotal for the successful clinical translation of DL-MRI methods. Future studies are warranted to rigorously analyze and interpret noise behaviour and to develop noise modeling strategies that can accomodate local image content—ultimately ensuring that diagnostic regions of interest remain free from deleterious noise artifacts.

The variance estimation framework presented in this work naturally aligns with the broader concept of \emph{uncertainty quantification} in deep neural networks and primarily captures \emph{aleatoric} uncertainty, arising from measurement noise in k-space. By interpreting voxel-wise variance as an indicator of low SNR or ill-conditioned regions, this approach facilitates reliable confidence assessments in clinical settings, helping guide diagnostic decisions (“Is this subtle lesion credible?”) and acquisition strategies (“Should additional k-space samples be acquired here?”). Our proposed estimator efficiently computes voxel-wise uncertainty via JVPs, making it a promising approach for large deep MRI reconstruction pipelines to explicitly characterize noise and enhance interpretability. In the future, this method could be combined with \emph{inherently stochastic reconstruction techniques}, i.e. diffusion models \cite{sure_mardani,Chung2022}, to also capture \emph{epistemic} uncertainty arising from the model. While our current focus is on estimating the \emph{diagonal} elements of the image covariance matrix (i.e., voxelwise variances), extending these techniques to capture \emph{off-diagonal} terms also remains an important frontier. For instance, in functional MRI, cross-covariance structures could illuminate shared noise sources or functional connectivity across different brain regions \cite{brain_covariance,fmri2}. 
% A unified framework that jointly models diagonal and off-diagonal entries would thus yield deeper insights into group-wise neuropatterns, artifact propagation, and inter-voxel dependencies. Developing efficient algorithms to achieve this, particularly for large volumes and high-dimensional reconstructions, stands as a compelling challenge for future research.

% The principles and methodologies developed in this study hold significant potential for extension beyond MRI reconstruction to other imaging problems and dense pixel prediction tasks. 
Many imaging modalities, such as CT and PET share similar challenges related to noise characterization, reconstruction under limited data, and estimating uncertainty in predictions \cite{CT}. For instance, in CT and PET, where dose reduction is a key objective, quantifying uncertainty due to measurement noise could enhance confidence in low-dose reconstructions, improving clinical decision-making \cite{low_dose_ct}.  Moreover, principles developed in this study hold significant potential for broader image-to-image tasks in computer vision, including super-resolution, denoising, inpainting, or synthesis by appropriate modifications based on the input noise distribution and forward corruption model \cite{ledig2017photo, zhang2017beyond,pathak2016context,resvit}. In these applications, accurately estimating the spatial distribution of uncertainty at the pixel level can improve interpretability and trustworthiness, especially in high-stakes scenarios such as medical diagnostics or autonomous systems \cite{kendall2017uncertainties}. 

In conclusion, we introduced a model-agnostic Jacobian sketching algorithm for estimating voxel-wise noise variance in DL MRI reconstructions. Our approach efficiently and accurately captures how acquisition noise propagates through nonlinear, multi-coil reconstructions without costly MC simulations. Extensive experiments demonstrate strong alignment with reference variance maps across diverse architectures, suggesting that our method can be readily incorporated into existing deep reconstruction pipelines. Beyond improving interpretability of DL reconstructions, these noise estimates hold promise for guiding adaptive acquisition, assessing image quality, and facilitating clinically actionable uncertainty quantification.

% Future extensions may also incorporate advanced techniques, such as probabilistic deep learning or Bayesian methods, to refine uncertainty quantification across diverse imaging domains \cite{gal2016dropout, kendall2017uncertainties}.

\newpage
\section*{Acknowledgements}
Authors would like to thank Moritz Alexander Bolling for helpful discussions on first order methods and statistical tests. This work was supported by the following grants: NIH R01AR077604, R01EB002524, R01AR079431, P41EB027060
\section*{Impact Statement}

This work aims to enhance the reliability and interpretability of deep learning-based MRI reconstruction methods, potentially improving clinical decision-making by providing more accurate and trustworthy imaging tools. While its primary objective is to advance the field of machine learning, we acknowledge that such advancements may have broader societal implications, including increased accessibility to high-quality diagnostic imaging and more efficient healthcare workflows. At present, we see no specific negative ethical or societal issues that warrant particular attention.

\nocite{langley00}

\bibliography{example_paper}
\bibliographystyle{icml2025}

%%%%%%%%%%%%%%%%%%%%%%%%%%%%%%%%%%%%%%%%%%%%%%%%%%%%%%%%%%%%%%%%%%%%%%%%%%%%%%%
%%%%%%%%%%%%%%%%%%%%%%%%%%%%%%%%%%%%%%%%%%%%%%%%%%%%%%%%%%%%%%%%%%%%%%%%%%%%%%%
% APPENDIX
%%%%%%%%%%%%%%%%%%%%%%%%%%%%%%%%%%%%%%%%%%%%%%%%%%%%%%%%%%%%%%%%%%%%%%%%%%%%%%%
%%%%%%%%%%%%%%%%%%%%%%%%%%%%%%%%%%%%%%%%%%%%%%%%%%%%%%%%%%%%%%%%%%%%%%%%%%%%%%%
\newpage
\appendix
\onecolumn
% \appendix
\section{Detailed Noise Model for Multi-Coil MRI}
\label{appendix:coil_noise_deriv}

\subsection{Coil Noise as Linear Combinations of Independent Sources}
\label{appendix:noise_linear_comb}
We assume that noise in each coil $\gamma$ arises from an ensemble of zero-mean independent Gaussian
sources $\{\xi_\tau\}$, each with standard deviation $\sigma_\tau$ \cite{macovski_noise}. Concretely,
\begin{equation}
\eta_{\gamma}(t)
\;=\;
\sum_{\tau}\;\omega_{(\gamma,\tau)}\,\xi_{\tau}(t).
\end{equation}
Because each $\xi_\tau$ is Gaussian, $\eta_{\gamma}(t)$ must also be Gaussian. The weight
$\omega_{(\gamma,\tau)}$ indicates how strongly the $\tau$-th noise source couples to coil
$\gamma$.

\subsection{Coil Covariance Matrix $\widetilde{\bm{\Sigma}}_{k}$}
Summarizing these noise contributions yields an $n_c\times n_c$ \emph{coil covariance matrix}
\(\widetilde{\bm{\Sigma}}_{k}\), where
\begin{equation}
\label{eq:coil_cov_block}
(\widetilde{\bm{\Sigma}}_{k})_{\gamma,\gamma'}
\;=\;
\sum_{\tau} \sigma_\tau^2\;\omega_{(\gamma,\tau)}\;\bigl(\omega_{(\gamma',\tau)}\bigr)^*.
\end{equation}
Here, $(\widetilde{\bm{\Sigma}}_{k})_{\gamma,\gamma}$ is the noise variance in coil $\gamma$, and 
off-diagonal elements encode coil-to-coil covariances, and $n_c$ is the number of channels in the imaging system.
Since noise originates from physical sources that induce real-valued fluctuations, the cross-correlations between coils must satisfy Hermitian symmetry due to Maxwellian reciprocity in electromagnetic induction \cite{roemer1990nmr, brown2014magnetic}. Consequently, $\widetilde{\bm{\Sigma}}_{k}$ is \emph{Hermitian} (\(\gamma,\gamma'\) swap with a conjugate to see mathematically) and \emph{positive semidefinite} by construction (as all covariance matrices), since it can be 
expressed in the form
\[
\mathbf{W}\,\mathbf{D}\,\mathbf{W}^H,
\]
where $\mathbf{W}$ contains $\omega_{(\gamma,\tau)}$, and $\mathbf{D}$ is diagonal of 
$\{\sigma_\tau^2\}$. Thus, for any vector $\bm{v}\in \mathbb{C}^{n_c}$,
\[
\bm{v}^H\,\widetilde{\bm{\Sigma}}_{k}\,\bm{v} 
\;=\;
\|\mathbf{D}^{1/2}\,\mathbf{W}^H\,\bm{v}\|^2 
\;\ge\; 0,
\]
confirming positive semidefiniteness.

\subsection{Block-Diagonal Sample Covariance \texorpdfstring{$\bm{\Sigma}_{k}$}{Sigma\_k}}

Suppose noise is uncorrelated across $n_f$ distinct \emph{k}-space frequency locations, while 
retaining a coil-to-coil covariance $\widetilde{\bm{\Sigma}}_{k}\in\mathbb{C}^{n_c\times n_c}$ 
at each frequency (where $n_c$ is the number of coils). If $\bm{n}\in\mathbb{C}^{m}$, with 
$m=n_f\cdot n_c$, denotes the stacked multi-coil noise over $n_f$ frequency bins, then the 
\emph{sample covariance matrix} of $\bm{n}$ can be expressed as a block diagonal:
\[
\bm{\Sigma}_{k} 
\;=\; 
\underbrace{\mathrm{diag}\!\Bigl(\widetilde{\bm{\Sigma}}_{k},\;\dots,\;\widetilde{\bm{\Sigma}}_{k}\Bigr)}_{n_f\text{ blocks}}
\;\in\;\mathbb{C}^{m\times m}.
\]
Equivalently, we may write
\[
\bm{\Sigma}_{k} 
\;=\;
\bm{I}_{n_f}
\,\otimes\,
\widetilde{\bm{\Sigma}}_{k},
\]
where $\bm{I}_{n_f}$ is the $n_f\times n_f$ identity matrix and $\otimes$ denotes the Kronecker product.  
Each $n_c\times n_c$ diagonal block $\widetilde{\bm{\Sigma}}_{k}$ captures the coil-to-coil noise correlation 
structure for one frequency index, and repeating these blocks $n_f$ times encodes independence across 
frequencies.

In practice, one can exploit a \emph{compact} representation in which $\widetilde{\bm{\Sigma}}_{k}$ 
is used directly during linear-algebraic factorizations or operator-based manipulations, 
rather than forming the full $m\times m$ block diagonal.  
Such a compact approach reduces both memory and computational overhead significantly, 
as the repeated blocks need not be explicitly stored or inverted. Consequently, we directly utilize the coil covariance matrix $\widetilde{\bm{\Sigma}}_{k}$ in our computational framework for efficiency.
Nevertheless, for theoretical analysis and proofs carried out in this study, 
it is more convenient to leverage the sample covariance matrix $\bm{\Sigma}_k$, 
thereby making explicit how noise correlations factor into the imaging operator $\bm{A}$ and 
the subsequent derivations.
We now show that if $\widetilde{\bm{\Sigma}}_{k}$ is Hermitian positive semidefinite (HPSD), then $\bm{\Sigma}_k$ inherits these properties.

\paragraph{Hermitian.}
A block-diagonal matrix is Hermitian if and only if each diagonal block is Hermitian.  
Since $\widetilde{\bm{\Sigma}}_k^H = \widetilde{\bm{\Sigma}}_k$, we have
\[
\bm{\Sigma}_k^H 
=
(\bm{I}_{n_f} \otimes \widetilde{\bm{\Sigma}}_k)^H
=
\bm{I}_{n_f}^H \otimes \widetilde{\bm{\Sigma}}_k^H
=
\bm{I}_{n_f}\otimes\widetilde{\bm{\Sigma}}_k
=
\bm{\Sigma}_k.
\]
Hence, $\bm{\Sigma}_k$ is Hermitian.

\paragraph{Positive Semidefinite (PSD).}
A Hermitian $\bm{\Sigma}_k$ is PSD if and only if $\mathbf{x}^H \bm{\Sigma}_k \mathbf{x} \ge 0$ 
for every $\mathbf{x}\in\mathbb{C}^m$. Partition $\mathbf{x}$ into $n_f$ sub-vectors, 
$\mathbf{x} = (\mathbf{x}_1,\dots,\mathbf{x}_{n_f})$, each $\mathbf{x}_i \in \mathbb{C}^{n_c}$.  
Because $\bm{\Sigma}_k$ is block diagonal, we get
\[
\mathbf{x}^H \,\bm{\Sigma}_k\,\mathbf{x}
=
\sum_{i=1}^{n_f}
\Bigl(
\mathbf{x}_i^H\,\widetilde{\bm{\Sigma}}_k\,\mathbf{x}_i
\Bigr).
\]
If $\widetilde{\bm{\Sigma}}_k$ is PSD, then each $\mathbf{x}_i^H \,\widetilde{\bm{\Sigma}}_k\, \mathbf{x}_i \ge 0$.  
Summing these terms yields a nonnegative result, thus $\bm{\Sigma}_k$ is PSD.

\section{Existence of the Network Jacobian}
\label{appendix:jacobian_existence}

Unrolled MRI networks alternate between 
\emph{regularization} and \emph{data-consistency} updates for $K$ iterations, yielding a final 
reconstruction $\bm{x}^{(K)}$ \cite{unrolled_1,unrolled_2}. Concretely, each iteration has two steps:

\begin{enumerate}

\item \textbf{Regularization:} 
  \[
    \hat{\bm{x}}^{(k)} 
    \;=\; 
    \mathcal{F}^{(k)}\!\Bigl(\bm{x}^{(k-1)}\Bigr),
  \]
  where $\mathcal{F}^{(k)}(\cdot)$ is a \emph{neural network} or learned operator (e.g., a CNN 
  block) that acts as a trainable regularizer.  Neural networks are well-known to be differentiable
  with respect to their inputs, assuming standard operations (convolution, ReLU, batch norm, etc.) 
  \cite{Goodfellow-et-al-2016}. Hence, the map 
  $\bm{x}^{(k-1)} \mapsto \hat{\bm{x}}^{(k)}$ is differentiable.  

\item \textbf{Data Consistency (DC):}
  \[
    \bm{x}^{(k)} 
    \;=\; 
    \hat{\bm{x}}^{(k)}
    \;-\;
    \bm{A}^{H} \Bigl(\bm{A}\,\hat{\bm{x}}^{(k)} - \bm{y}\Bigr).
  \]
  Here, $\bm{A}\in \mathbb{C}^{m\times n}$ is the linear imaging operator, and $\bm{y}\in\mathbb{C}^{m}$ 
  is the measured data.  Since matrix-vector multiplication and addition/subtraction are 
  \emph{linear} (and thus differentiable) operations, the map 
  $\hat{\bm{x}}^{(k)} \mapsto \bm{x}^{(k)}$ is also differentiable.
\end{enumerate}

\paragraph{Composing Differentiable Steps.}
Because each iteration \emph{composes} two differentiable transformations, the combined $k$-th 
iteration step 
\[
  \bm{x}^{(k-1)} 
  \;\mapsto\;
  \bm{x}^{(k)}
\]
is itself differentiable. Consequently, unrolling $K$ times yields a chain of such transformations,
implying the final reconstruction $\bm{x}^{(K)}$ is a differentiable function of the initial 
estimate $\bm{x}^{(0)}$. Formally, 
\[
  f:
  \bm{x}^{(0)}
  \;\mapsto\;
  \bm{x}^{(K)}
  \;=\;
  \bigl(\underbrace{DC \circ \mathcal{F}^{(k)}\circ \cdots \circ DC \circ \mathcal{F}^{(1)}}_{\text{$K$ times}}\bigr)(\bm{x}^{(0)}).
\]
By the chain rule in multivariate calculus, this implies that $f$ has a well-defined Jacobian
$\bm{J}_f\!\in\!\mathbb{C}^{n\times n}$ at each input $\bm{x}^{(0)}$.

\begin{algorithm}[tb]
   \caption{Naive Per-Voxel Variance Calculation (Row-by-Row Jacobian)}
   \label{algo:naive_calc}
\begin{algorithmic}
   \REQUIRE 
   \begin{itemize}
      \item Reconstruction function $f: \mathbb{C}^{n} \to \mathbb{C}^{n}$ with Jacobian $\bm{J}_f \in \mathbb{C}^{n \times n}$,
      \item Imaging operator $\bm{A} \in \mathbb{C}^{m \times n}$,
      \item k-space sample covariance matrix $\bm{\Sigma}_k \in \mathbb{C}^{m \times m}$.
   \end{itemize}
   \ENSURE Voxel-wise variance map $\bigl\{\mathrm{Var}(\bm{x}_i)\bigr\}_{i=1}^n$. 

   \STATE \textbf{Compute the Cholesky factor of} $\bm{\Sigma}_k$:
   \[
       \bm{\Sigma}_k = \bm{\sigma}_k\,\bm{\sigma}_k^{H}.
   \]
   \STATE \textbf{Initialize an empty array for variances:}
   \[
       \mathrm{Var}(\bm{x}) = \bm{0} \in \mathbb{R}^{n}.
   \]

   \FOR{each voxel $i \;=\;1,\dots,n$}
       \STATE \textbf{Create one-hot vector} $\mathbf{e}_i \in \mathbb{R}^n$ with $1$ at index $i$, $0$ otherwise.
       \STATE \textbf{Backpropagate} $\mathbf{e}_i$ through $f$ to get row $i$ of $\bm{J}_f$:
       \[
         \nabla^\top f_i 
         \;\gets\; 
         \bigl(\bm{J}_f^H \,\mathbf{e}_i\bigr)^\top.
       \]
       \STATE \textbf{Form} 
       \[
         \bm{l}_i
         \;=\;
         (\nabla^\top f_i)\;\bm{A}^H\;\bm{\sigma}_k.
       \]
       \COMMENT{Imaging operator and noise correlations included.}

       \STATE \textbf{Compute variance:}
       \[
         \mathrm{Var}(\bm{x}_i) 
         \;=\;
         \|\bm{l}_i\|_2^2.
       \]
       \STATE \textbf{Store in} $\mathrm{Var}(\bm{x})_i \gets \mathrm{Var}(\bm{x}_i)$.
   \ENDFOR

\end{algorithmic}
\end{algorithm}

\section{Verification of Random Vector Properties}
\label{appendix:random_vectors}
Here, we first demonstrate that the introduced random vectors—complex Gaussian and Rademacher—satisfies the conditions stipulated in Theorem~\ref{thm:unbiased_estimator}: zero mean and unit covariance.

\subsection{Complex Gaussian Vectors}

\begin{definition}
A \textbf{Complex Gaussian vector} \(\bm{v} \in \mathbb{C}^{n}\) is defined such that each element \(v_j\) is independently sampled from a complex normal distribution with zero mean and unit variance, denoted as \(v_j \sim \mathcal{CN}(0, 1)\). This implies:
\[
v_j = a_j + i b_j,
\]
where \(a_j\) and \(b_j\) are independent real-valued Gaussian random variables with \(a_j, b_j \sim \mathcal{N}(0, 0.5)\).
\end{definition}

\subsubsection{Zero Mean}

The mean of each element \(v_j\) is calculated as:
\[
\mathbb{E}[v_j] = \mathbb{E}[a_j] + i \mathbb{E}[b_j] = 0 + i \cdot 0 = 0
\]
Therefore,
\[
\mathbb{E}[\bm{v}] = \bm{0}
\]

\subsubsection{Unit Covariance}

Since \(a_j\) and \(b_j\) are independent and each has a variance of \(0.5\), the covariance between \(v_j\) and \(v_k\) is:
\[
\mathbb{E}[v_j v_k^{H}] = \mathbb{E}[(a_j + i b_j)(a_k - i b_k)] = \mathbb{E}[a_j a_k] + \mathbb{E}[b_j b_k]
\]
Given that \(a_j\) and \(b_j\) are independent across different indices:
\[
\mathbb{E}[v_j v_k^{H}] = 
\begin{cases}
1, & \text{if } j = k, \\
0, & \text{if } j \neq k.
\end{cases}
\]
Thus,
\[
\mathbb{E}[\bm{v} \bm{v}^{H}] = \bm{I}
\]

\subsection{Complex Rademacher Vectors}

\begin{definition}
A \textbf{Complex Rademacher vector} \(\bm{v} \in \mathbb{C}^{n}\) is defined such that each element \(v_j\) is independently sampled from the set \(\{e^{j \theta} \mid \theta \sim \text{Uniform}[0, 2\pi)\}\). In other words, each \(v_j\) is of the form:
\[
v_j = e^{j \theta_j},
\]
where \(\theta_j\) is uniformly distributed over the interval \([0, 2\pi)\).
\end{definition}

\subsubsection{Zero Mean}

Since \(\theta_j\) is uniformly distributed over \([0, 2\pi)\), the expectation of each \(v_j\) is:
\[
\mathbb{E}[v_j] = \mathbb{E}[e^{j \theta_j}] = \frac{1}{2\pi} \int_0^{2\pi} e^{j \theta} d\theta = 0
\]
Therefore,
\[
\mathbb{E}[\bm{v}] = \bm{0}
\]

\subsubsection{Unit Covariance}

For \(j = k\),
\[
\mathbb{E}[v_j v_j^{H}] = \mathbb{E}[e^{j \theta_j} e^{-j \theta_j}] = \mathbb{E}[1] = 1
\]
For \(j \neq k\),
\[
\mathbb{E}[v_j v_k^{H}] = \mathbb{E}[e^{j \theta_j} e^{-j \theta_k}] = \mathbb{E}[e^{j \theta_j}] \cdot \mathbb{E}[e^{-j \theta_k}] = 0 \cdot 0 = 0
\]
This separation of expectations is valid due to the independence of \(\theta_j\) and \(\theta_k\). Hence,
\[
\mathbb{E}[\bm{v} \bm{v}^{H}] = \bm{I}
\]
\section{Variance of Complex Gaussian vs. Complex Random-Phase}
\label{appendix:complex_variance}

Let \(\bm{\Sigma}\in \mathbb{C}^{n\times n}\) be a Hermitian positive
semi-definite matrix (e.g., a covariance matrix), and let \(\bm{v}\in \mathbb{C}^n\) be a
random vector satisfying
\[
  \mathbb{E}[\bm{v}] 
  \;=\;
  \bm{0},
  \quad
  \mathbb{E}\bigl[\bm{v}\,\bm{v}^{H}\bigr]
  \;=\;
  \bm{I}.
\]
We define the following unbiased estimator for the diagonal entry \(\Sigma_{ii}\):
\[
  Y_i 
  \;=\; 
  \bigl(\bm{\Sigma}\,\bm{v}\bigr)_i \,\bigl(v_i\bigr)^{*},
  \quad\Longrightarrow\quad
  \mathbb{E}[Y_i] 
  \;=\; 
  \Sigma_{ii}.
\]
Below, we derive the variance of \(Y_i\) under two complex random-vector distributions:
\emph{(i)~complex Gaussian} and \emph{(ii)~complex random-phase (Rademacher)}. 
In both cases, the key difference lies in the \emph{fourth moment} of the entries.

\subsection{General Expansion of \(\lvert Y_i\rvert^2\)}

First, we write out
\[
  Y_i 
  \;=\; 
  \sum_{j=1}^n \Sigma_{ij}\,v_j
  \;\times\;
  \bigl(v_i\bigr)^{*}.
\]
Hence,
\[
  \lvert Y_i\rvert^2
  \;=\;
  \Bigl(\sum_{j=1}^n \Sigma_{ij}\,v_j\Bigr)\,\bigl(v_i\bigr)^{*}
  \;\times\;
  \Bigl(\sum_{k=1}^n \Sigma_{ik}\,v_k\Bigr)^{*}\,\bigl(v_i\bigr)^{\!**}.
\]
Since \(\bm{\Sigma}\) is Hermitian, \(\Sigma_{ik}^{*} = \Sigma_{ki}\), and
\((v_i)^{**} = v_i\).  Expanding yields
\[
  \lvert Y_i\rvert^2
  \;=\;
  \Bigl(\sum_{j=1}^n \Sigma_{ij}\,v_j\Bigr)\,
  \Bigl(\sum_{k=1}^n \Sigma_{ik}\,v_k\Bigr)^{\!*}
  \;\times\;
  \lvert v_i\rvert^2.
\]
Taking expectation,
\[
  \mathbb{E}\bigl[\lvert Y_i\rvert^2\bigr]
  \;=\;
  \sum_{j=1}^n \sum_{k=1}^n \Sigma_{ij}\,\Sigma_{ik}^{*}
  \;\mathbb{E}\bigl[v_j\,v_k^{*}\,\lvert v_i\rvert^2\bigr].
\]
Because \(\mathbb{E}[\bm{v}\,\bm{v}^{H}] = \bm{I}\), we know
\(\mathbb{E}[v_j\,v_k^{*}]=\delta_{jk}\).  For \(j\neq i\), \(k\neq i\), the 
entries \(v_j\) and \(v_i\) are uncorrelated in magnitude, so 
\(\mathbb{E}[\,\lvert v_i\rvert^2\,v_j\,v_k^{*}\bigr] 
= \mathbb{E}[\lvert v_i\rvert^2]\,\delta_{jk} = 1\cdot\delta_{jk}\).
When \(j=i\) or \(k=i\), we retain the \emph{fourth moment} term
\(\mathbb{E}[\,\lvert v_i\rvert^4\,]\). 
Thus, generically,
\[
  \mathbb{E}\bigl[\lvert Y_i\rvert^2\bigr]
  \;=\;
  \sum_{j\neq i} \lvert\Sigma_{ij}\rvert^2\;\mathbb{E}\bigl[\lvert v_j\rvert^2\,\lvert v_i\rvert^2\bigr]
  \;+\;
  \lvert \Sigma_{ii}\rvert^2\,\mathbb{E}\bigl[\lvert v_i\rvert^4\bigr].
\]
Given \(\mathbb{E}\bigl[\lvert v_j\rvert^2\bigr] = 1\) for all \(j\), we
ultimately get
\[
  \mathbb{E}\bigl[\lvert Y_i\rvert^2\bigr]
  \;=\;
  \sum_{j \neq i}\lvert \Sigma_{ij}\rvert^2
  \;+\;
  \lvert \Sigma_{ii}\rvert^2\,\mathbb{E}\bigl[\lvert v_i\rvert^4\bigr].
\]
Subtracting \(\lvert\Sigma_{ii}\rvert^2\) (which is
\(\bigl\lvert \mathbb{E}[Y_i]\bigr\rvert^2\)) then gives the variance:
\[
  \mathrm{Var}[Y_i]
  \;=\;
  \Bigl[
    \sum_{j \neq i}\lvert \Sigma_{ij}\rvert^2
    \;+\;
    \lvert \Sigma_{ii}\rvert^2\,\mathbb{E}\bigl[\lvert v_i\rvert^4\bigr]
  \Bigr]
  \;-\;
  \lvert \Sigma_{ii}\rvert^2.
\]
Thus the crux is: \(\mathbb{E}[\lvert v_i\rvert^4]\).

\subsection{Complex Gaussian Case}

\paragraph{Distribution.}
Let each \(v_i\) be drawn from a circularly symmetric Gaussian 
(\(\mathcal{CN}(0,1)\)), implying
\[
  \mathbb{E}\bigl[\lvert v_i\rvert^2\bigr] = 1,
  \quad
  \mathbb{E}\bigl[\lvert v_i\rvert^4\bigr] = 2.
\]
(That is, if \(v_i = \frac{1}{\sqrt{2}}(x_i + j\,y_i)\) with 
\(x_i,y_i \sim \mathcal{N}(0,1)\) i.i.d.)

\paragraph{Variance Calculation.}
Substitute \(\mathbb{E}[|v_i|^4]=2\) into the generic formula:
\[
  \mathbb{E}\bigl[\lvert Y_i\rvert^2\bigr]
  \;=\;
  \sum_{j \neq i}\lvert\Sigma_{ij}\rvert^2
  \;+\;
  2\,\lvert\Sigma_{ii}\rvert^2.
\]
Therefore,
\[
  \mathrm{Var}[Y_i]
  \;=\;
  \sum_{j \neq i}\lvert\Sigma_{ij}\rvert^2
  \;+\;
  2\,\lvert\Sigma_{ii}\rvert^2
  \;-\;
  \lvert\Sigma_{ii}\rvert^2
  \;=\;
  \sum_{j=1}^n \lvert\Sigma_{ij}\rvert^2.
\]
Hence, for \emph{complex Gaussian}:
\[
  \mathrm{Var}[Y_i]
  \;=\;
  \sum_{j=1}^n \bigl\lvert\Sigma_{ij}\bigr\rvert^2.
\]

\subsection{Complex Random-Phase (Rademacher) Case}

\paragraph{Distribution.}
Now let each \(v_i\) have \emph{unit magnitude} and uniformly random phase,
\[
  \lvert v_i\rvert = 1,
  \quad
  v_i \;=\; e^{j\,\theta_i},
  \quad
  \theta_i \sim \mathrm{Uniform}[0,2\pi].
\]
Then
\[
  \mathbb{E}\bigl[\lvert v_i\rvert^4\bigr]
  \;=\;
  1.
\]

\paragraph{Variance Calculation.}
Substituting \(\mathbb{E}[|v_i|^4]=1\) into our generic formula:
\[
  \mathbb{E}\bigl[\lvert Y_i\rvert^2\bigr]
  \;=\;
  \sum_{j \neq i}\lvert\Sigma_{ij}\rvert^2
  \;+\;
  1\cdot \lvert\Sigma_{ii}\rvert^2
  \;=\;
  \sum_{j=1}^n \lvert\Sigma_{ij}\rvert^2.
\]
Subtracting \(\lvert\Sigma_{ii}\rvert^2\) from this, we get
\[
  \mathrm{Var}[Y_i]
  \;=\;
  \sum_{j=1}^n \lvert\Sigma_{ij}\rvert^2
  \;-\;
  \lvert\Sigma_{ii}\rvert^2.
\]
Thus, for \emph{complex random-phase} (or ``complex Rademacher''):
\[
  \mathrm{Var}[Y_i]
  \;=\;
  \sum_{j=1}^n \lvert\Sigma_{ij}\rvert^2 
  \;-\;
  \lvert \Sigma_{ii}\rvert^2.
\]

\begin{algorithm}[tb]
   \caption{Noise Calculation in DL-based MRI reconstruction via Sketching the Network Jacobian}
   \label{alg:variance_estimation_matrix}
\begin{algorithmic}
   \REQUIRE
   \begin{itemize}
      \item Imaging operator $\bm{A} \in \mathbb{C}^{m \times n}$,
      \item k-space sample covariance matrix $\bm{\Sigma}_k \in \mathbb{C}^{m \times m}$
      \item Unrolled recon $f : \mathbb{C}^n \to \mathbb{C}^n$ with Jacobian $\bm{J}_f$,
      \item $S$, size of random matrix $\bm{V}_S$.
   \end{itemize}
   \ENSURE Approximate $\widehat{\mathrm{diag}\,\bigl(\bm{\Sigma}_{\bm{x}}\bigr)} \approx\mathrm{diag}\bigl(\bm{\Sigma}_{\bm{x}}\bigr)$, where 
   $\bm{\Sigma}_{\bm{x}} = \bm{J}_f\,\bm{A}^H\,\bm{\Sigma}_k\,\bm{A}\,\bm{J}_f^H$

   \STATE Compute Cholesky factorization: $\bm{\Sigma}_k = \bm{\sigma}_k\bm{\sigma}_k^H$

   \STATE $\bm{V}_S \in \mathbb{C}^{m\times S}$ with columns $\{\bm{v}_j\}$ sampled from a \emph{complex random-phase} distribution:
   \[
     v_{i}^{(j)} 
     \;=\; 
     e^{\,j\,\theta_{i}^{(j)}},
     \quad
     \theta_{i}^{(j)} \sim \mathrm{Uniform}[0,2\pi].
   \]
   \STATE $\bm{W}_S \gets \bm{\sigma}_k \,\bm{V}_S \quad(\in \mathbb{C}^{m\times S})$
   \STATE $\widetilde{\bm{W}}_S \gets \bm{A}^{H}\,\bm{W}_S \quad(\in \mathbb{C}^{n\times S})$
   \STATE $\bm{U}_S \gets \bm{J}_f\bigl(\widetilde{\bm{W}}_S\bigr)$ 
   \STATE $\bm{V}_{\text{samples}} \gets \bm{U}_S \odot \bm{U}_S^H \quad(\in \mathbb{R}^{n\times S})$
   \STATE $\widehat{\mathrm{diag}\,\bigl(\bm{\Sigma}_{\bm{x}}\bigr)} 
          \;\gets\; 
          \frac{1}{S}\,\bm{V}_{\text{samples}}\,\mathbf{1}_S
          \quad(\in \mathbb{R}^n)$
\end{algorithmic}
\end{algorithm}

\section{Sketching Matrix}
\label{appendix:sketching_matrix}
\begin{figure*}[ht]
    \centering
    \includegraphics[width=0.9\columnwidth]{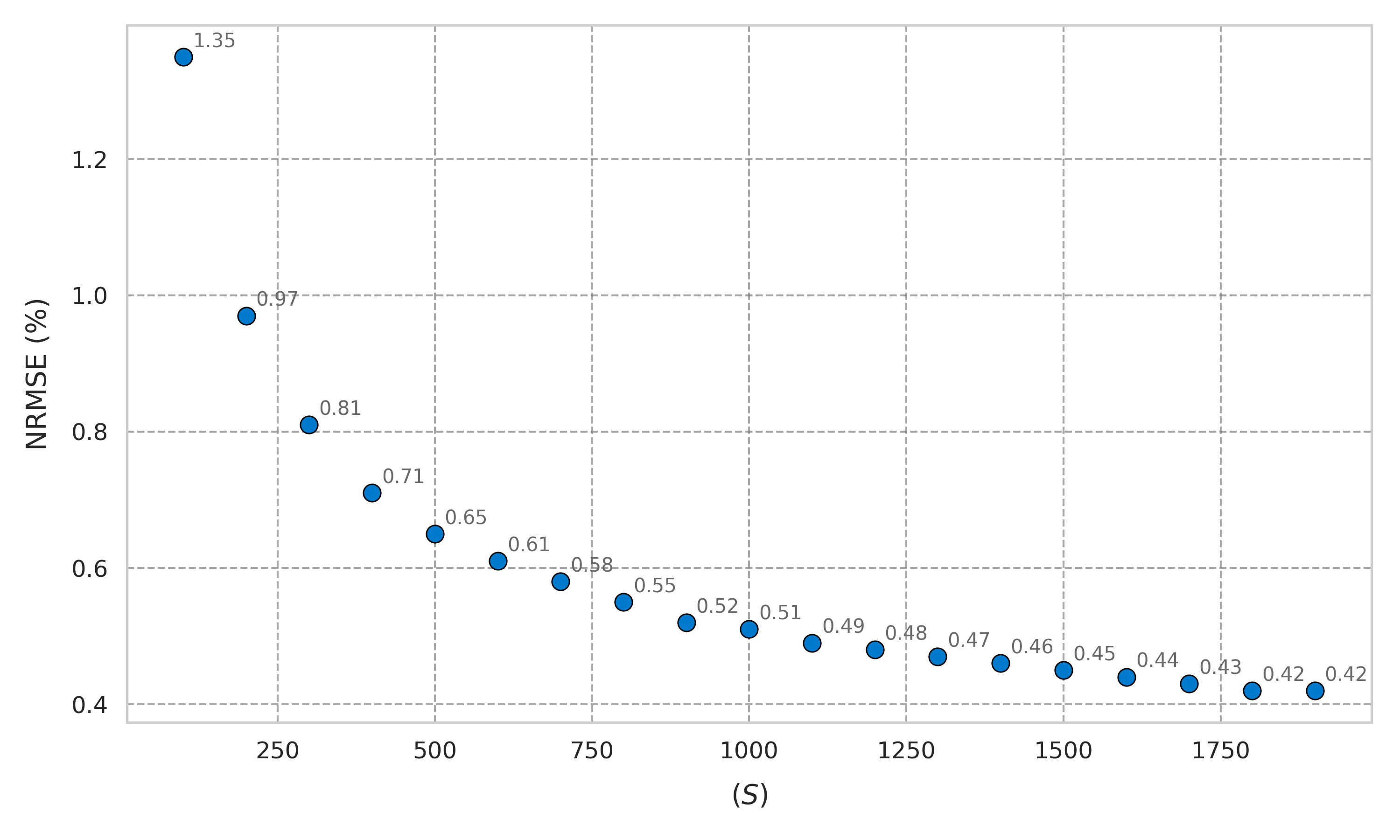}
    \caption{
    NRMSE of the proposed method over test slices vs.\ size of the sketching matrix ($S$) (E2E-VarNet on brain data at $R=8, \alpha=1
$).
    As the size $S$ of the sketching matrix $\bm{V}_S$
    grows from 100 to 1,900, the Normalized Root-Mean-Square Error (NRMSE) of the variance
    estimator decreases. While additional probing vectors generally improve accuracy by
    capturing more of the Jacobian’s structure, the marginal benefit tapers off beyond about
    1,000--1,200 vectors, suggesting a practical trade-off between improved precision and
    increased computation time.
    }
    \label{fig:nrmse_vs_samples}
\end{figure*}

Figure~\ref{fig:nrmse_vs_samples} varies $S$, size of the sketching matrix $\bm{V}_S$, and measures the resulting NRMSE of our variance estimator. As $S$ increases from 100 to 1,900, there is a subtle decrease in error. This trend reflects the improved accuracy of the diagonal estimator when more random vectors probe the Jacobian structure. However, beyond a certain point (approximately $S=1,000$--$1,200$ in our experiments), the NRMSE decreases more gradually, indicating diminishing returns in accuracy. Thus, while a large $S$ can yield more precise variance maps, it also incurs higher computational cost; our choice of $S=1,000$ would yield a favorable balance between accuracy and runtime for most practical settings.

\section{Empirical Noise Calculation}
\label{appendix:monte}

\begin{figure*}[ht]
    \vskip 0.2in
    \begin{center}
    \centerline{\includegraphics[width=\columnwidth]{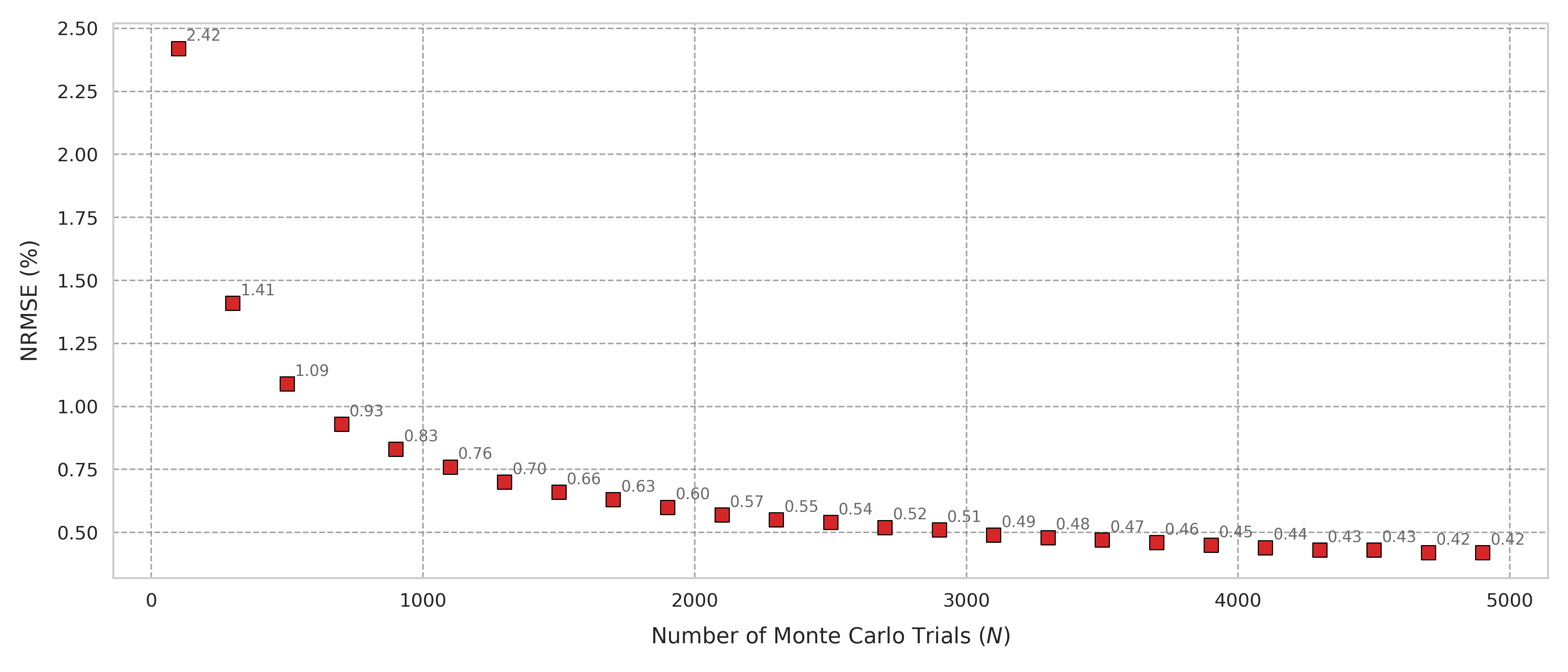}}
    \caption{
    Mean NRMSE of empirical baseline over test slices vs.\ the number of MC Trials (N) (E2E-VarNet on brain data at $R=8,\alpha=1$). 
    We vary the total number of MC realizations from 100 up to 5,000
    and compare the resulting variance maps to a high-sample reference 
    (e.g., 10,000 trials). As more trials are included, the empirical variance 
    estimation converges around $N=3000$, shown here by the decreasing NRMSE. 
    }
    \label{fig:nrmse_vs_N_monte_carlo}
    \end{center}
    \vskip -0.2in
\end{figure*}

\paragraph{MC Trials.}
For each experiment (reconstruction method, noise level, acceleration rate, dataset) and slice, we simulated
$N=3,000$ MC trials. Specifically, we draw the noise vector $\bm{n}$ from
$\bm{\Sigma}_k$ for each trial, add it to the measured \emph{k}-space, then reconstruct
the image. Across these 3,000 reconstructions, we compute voxelwise sample variance to
form an \emph{empirical} variance map, which serves as the “gold standard” reference. 
\paragraph{Convergence.}
Although we typically fix $3{,}000$ trials to ensure a stable reference variance, 
Figure~\ref{fig:nrmse_vs_N_monte_carlo} illustrates the effect of varying the trial count 
from $100$ to $5{,}000$.  For smaller trial counts, the empirical variance estimate shows 
notably larger NRMSE relative to a high-sample baseline (e.g., $10{,}000$ trials), but this 
error diminishes as we approach $2{,}500$--$3{,}000$ trials, thereafter offering only 
gradual improvements. Consequently, while increasing the number of MC samples 
monotonically improves estimation accuracy, it quickly becomes computationally prohibitive to 
reconstruct each perturbed k-space many thousands of times.

\section{Reconstruction Methods}
\label{appendix:recon_methods}
Here we provide detailed descriptions of the six deep MRI reconstruction algorithms evaluated in our study. Each algorithm falls under a distinct combination of \textit{learning paradigm} and \textit{architecture design}, ranging from supervised to self-supervised, and from fully data-driven to physics-driven unrolled models. All neural architectures incorporate complex-valued inputs via stacking real and imaginary parts,  resulting in 2 input and output channels. Note that hyperparameters for each method were tuned on the validation dataset to ensure optimal performance. Unless otherwise stated, all models were trained via Adam optimizer \cite{adam}, with $\beta_1=0.965,\beta_2=0.99$. Here, we provide detailed descriptions for each method. We used PyTorch for modeling \cite{pytorch}, and NVIDIA NVIDIA RTX A6000 for GPU acceleration.
\subsection{Supervised, Physics-Driven Unrolled}

\subsubsection{E2E-VarNet \cite{e2e_varnet}}
\label{appendix:e2e-varnet}
\textbf{E2E-VarNet} is a multi-coil unrolled reconstruction method that alternates between 
learned regularization blocks and data-consistency (DC) steps. Each iteration applies a 
trainable neural network (often a ResNet or similar CNN) to refine the current image estimate, 
followed by a DC step enforcing agreement with the measured multi-coil \emph{k}-space data.

We configure E2E-VarNet with 4 unrolled steps, each containing 2 ResNet blocks (256 channels, 
kernel size 3). An $\ell_1$-norm reconstruction loss is used, and the network is optimized 
with Adam~(lr=$10^{-4}$). Table~\ref{tab:e2e_varnet_configs} summarizes additional 
hyperparameters for knee and brain MRI datasets.

\begin{table}[!ht]
\centering
\small
\caption{
Key Hyperparameters used for training E2E-VarNet on knee and brain datasets.
}
\label{tab:e2e_varnet_configs}
\begin{tabular}{lcc}
\toprule
\textbf{Hyperparameter}          & \textbf{Knee}     & \textbf{Brain} \\
\midrule
\multicolumn{3}{l}{\emph{Architecture}} \\[3pt]
Unrolled Steps                   & 4                 & 4                     \\
Block Architecture               & ResNet            & ResNet                \\
ResNet Channels                  & 256               & 256                   \\
\# ResNet Blocks / Step          & 2                 & 2                     \\
Kernel Size                      & 3                 & 3                     \\
\midrule
 \emph{Training}& &\\
Loss Function                    & $\ell_1$          & $\ell_1$              \\
\textbf{Batch Size}              & 4                 & 4                     \\
\textbf{Grad. Accum. Steps}      & 4                 & 4                     \\
Base Learning Rate               & $10^{-4}$         & $10^{-4}$             \\
Weight Decay                     & $10^{-4}$         & $10^{-4}$             \\
Max Iterations                   & 120{,}000         & 50{,}000              \\
\bottomrule
\end{tabular}
\end{table}

\subsubsection{MoDL \cite{modl}}
\label{appendix:modl}
\textbf{MoDL} (Model-Based Deep Learning) integrates conjugate-gradient (CG) steps for 
data-consistency with a CNN denoiser block for learned regularization. Each unrolled iteration 
iteratively solves a least-squares subproblem via CG and then applies a trainable ResNet 
denoiser.

We set MoDL to 4 unrolled iterations, each containing 10 internal CG steps. A ResNet 
(256 channels, 2 blocks per iteration) serves as the learned regularizer. Table~\ref{tab:modl_configs} 
details these hyperparameters, including $\ell_1$ reconstruction loss, Adam (lr=$10^{-4}$), 
and separate CG tolerances for knee vs.\ brain.

\begin{table}[!ht]
\centering
\small
\caption{
Key Hyperparameters used for training MoDL on knee and brain datasets.
}
\label{tab:modl_configs}
\begin{tabular}{lcc}
\toprule
\textbf{Hyperparameter}          & \textbf{Knee}              & \textbf{Brain}             \\
\midrule
\multicolumn{3}{l}{\emph{Architecture}} \\[3pt]
Unrolled Steps                   & 4                           & 4                          \\
DC Solver                        & CG (10 iterations, $\epsilon=10^{-5}$)& CG (10 iterations, $\epsilon=10^{-4}$)\\
Block Architecture               & ResNet            & ResNet                \\
 ResNet Channels                  & 256               &256                   \\
\# ResNet Blocks / Step          & 2                           & 2                          \\
Kernel Size                      & 3                           & 3                          \\
\bottomrule
 \emph{Training}& &\\
Loss Function                    & $\ell_1$                    & $\ell_1$                   \\
\textbf{Batch Size}              & 2                           & 2                          \\
\textbf{Grad. Accum. Steps}      & 8                           & 4                          \\
Base Learning Rate               & $10^{-4}$                   & $10^{-4}$                  \\
Weight Decay                     & $10^{-4}$                   & $10^{-4}$                  \\
Max Iterations                   & 80{,}000                    & 80{,}000                   \\
\bottomrule
\end{tabular}
\end{table}

\subsection{Supervised, Data-Driven}

\subsubsection{U-Net \cite{unet}}
\label{appendix:unet}

Unlike unrolled methods that tightly integrate the MRI forward model with iterative DC, 
\textbf{U-Net} is a fully data-driven, “one-shot” approach, mapping the zero-filled (or 
coil-combined) image directly to the reconstructed output. This yields faster inference, but it lacks physics-based constraints.

We use a standard U-Net with 4 levels of down/up-sampling, 32 base channels, and blocks of 
\texttt{\{conv, relu, conv, relu, batchnorm, dropout\}}. An $\ell_1$ reconstruction loss and 
Adam (lr=$10^{-3}$) train the network. Table~\ref{tab:unet_configs} summarizes key parameters 
for knee and brain datasets.

\begin{table}[!ht]
\centering
\small
\caption{Key Hyperparameters used for training U-Net on knee and brain datasets.}
\label{tab:unet_configs}
\begin{tabular}{lcc}
\toprule
\textbf{Hyperparameter}      & \textbf{Knee}        & \textbf{Brain} \\
\midrule
\multicolumn{3}{l}{\emph{Architecture}} \\[3pt]
Base Channels                & 32                   & 32                      \\
Depth (Pool Layers)          & 4                    & 4                       \\
Block Sequence               & \{conv, relu, conv, relu, bn, dropout\}
                             & \{conv, relu, conv, relu, bn, dropout\} \\
\bottomrule
 \emph{Training}& &\\
 Loss Function                & $\ell_1$             & $\ell_1$                \\
\textbf{Batch Size}          & 16                   & 12                      \\
\textbf{Grad. Accum. Steps}  & 1                    & 2                       \\
Base Learning Rate           & $10^{-3}$            & $10^{-3}$               \\
Weight Decay                 & $10^{-4}$            & $10^{-4}$               \\
Training Length              & 200 epochs           & 80{,}000 iterations     \\
\end{tabular}
\end{table}

\subsection{Semi-Supervised, Physics-Driven Unrolled}

\subsubsection{N2R (Noise2Recon) \cite{n2r}}
\label{appendix:n2r}
\textbf{N2R} leverages both a limited set of fully sampled ground-truth data and a larger set 
of undersampled data to train an unrolled network. Each iteration alternates between DC blocks 
and learned ResNet-based regularization blocks, enforcing consistency only on sampled \emph{k}-space 
points. 
Table~\ref{tab:n2r_configs_combined} outlines key hyperparameters, including the 
\emph{k}-space $\ell_1$ loss, 4 unrolled steps, 256-channel ResNet blocks, and consistency 
noise levels ([0.2, 0.5]) .

\begin{table}[!ht]
\centering
\small
\caption{Key hyperparameters used for training N2R on knee and brain datasets.
}
\label{tab:n2r_configs_combined}
\begin{tabular}{lcc}
\toprule
\textbf{Hyperparameter}       & \textbf{Knee}          & \textbf{Brain} \\
\midrule
\multicolumn{3}{l}{\emph{Architecture}} \\[3pt]
Total Training Scans          & 14 (13 undersamp.)     & 54 (50 undersamp.)   \\
Acceleration Factor           & 8                      & 8                    \\
Unrolled Steps                & 4                      & 4                    \\
Block Architecture            & ResNet (256 ch)        & ResNet (256 ch)      \\
\# ResNet Blocks / Step       & 2                      & 2                    \\
Kernel Size                   & 3                      & 3                    \\
\midrule
\multicolumn{3}{l}{\emph{Semi-Supervision}} \\[3pt]
Consistency noise std $\sigma$ Range& [0.2, 0.5]             & [0.2, 0.5]            \\
\midrule
\multicolumn{3}{l}{\emph{Training}} \\[3pt]
 Loss Function                 & \emph{k}-space $\ell_1$ &\emph{k}-space $\ell_1$ \\
\textbf{Batch Size}           & 2                      & 1                     \\
\textbf{Grad. Accum. Steps}   & 8                      & 4                     \\
Base Learning Rate            & $10^{-4}$              & $10^{-4}$             \\
Weight Decay                  & $10^{-4}$              & $10^{-4}$             \\
Max Iterations                & 80{,}000               & 80{,}000              \\
\bottomrule
\end{tabular}
\end{table}

\subsubsection{VORTEX \cite{vortex}}
\label{appendix:vortex}
\textbf{VORTEX} is another semi-supervised unrolled method, leveraging partial fully-sampled 
references alongside motion/noise augmentations. Each iteration incorporates DC steps and 
256-channel ResNet blocks. Random motion/noise transformations align with MR physics, 
enforcing consistent reconstructions under both fully and partially sampled data.

Like N2R, VORTEX employs 4 unrolled steps (2~ResNet blocks/step, 256 channels, kernel size=3). 
Loss is a mixture of $\ell_1$ and $\ell_2$ in the \emph{k}-space domain. Table~\ref{tab:vortex_configs} 
summarizes these hyperparameters, showing how motion/noise augmentations are tuned slightly differently between knee vs.\ brain data.

\begin{table}[!ht]
\centering
\small
\caption{Key hyperparameters used for training VORTEX on knee and brain datasets.
}
\label{tab:vortex_configs}
\begin{tabular}{lcc}
\toprule
\textbf{Hyperparameter}         & \textbf{Knee}          & \textbf{Brain} \\
\midrule
\multicolumn{3}{l}{\emph{Architecture}} \\[3pt]
Acceleration Factor             & 8                      & 8                     \\
Total Training Scans            & 14 (13 undersamp.)     & 54 (50 undersamp.)    \\
Unrolled Steps                  & 4                      & 4                     \\
Block Architecture              & ResNet (256 ch)        & ResNet (256 ch)       \\
\# ResNet Blocks / Step         & 2                      & 2                     \\
Kernel Size                     & 3                      & 3                     \\
\midrule
\multicolumn{3}{l}{\emph{VORTEX Motion/Noise Augmentations}} \\[3pt]
Motion Severity ($\alpha$) Range& [0.2, 0.5]            & [0.1, 0.2]  \\
Noise STD ($\sigma$) Range& [0.2, 0.5]            & [0.1, 0.2]  \\
\bottomrule
 \emph{Training}& &\\
 Loss Function                   & \emph{k}-space $\ell_1 / \ell_2$
                                & \emph{k}-space $\ell_1 / \ell_2$ \\
\textbf{Batch Size}             & 1                      & 1                     \\
\textbf{Grad. Accum. Steps}     & 4                      & 4                     \\
Base Learning Rate              & $10^{-4}$              & $10^{-4}$             \\
Weight Decay                    & $10^{-4}$              & $10^{-4}$             \\
Max Iterations                  & 80{,}000               & 80{,}000              \\
\end{tabular}
\end{table}

\subsubsection{SSDU \cite{ssdu}}
\label{appendix:ssdu}
\textbf{Self-Supervised Learning via Data Undersampling (SSDU)} obviates the need for fully 
sampled targets by splitting acquired \emph{k}-space data into “label” and “mask” sets. 
The “label” portion defines the reconstruction loss, while the “mask” portion is used within 
DC steps during the unrolled network’s forward pass. Mask ratio $\rho=0.2$ partitions 20\% 
of acquired samples as “label,” with the remaining 80\% guiding DC.

SSDU’s unrolled network matches E2E-VarNet’s structure (4 steps, ResNet blocks 
at 256 channels). All training references were undersampled, partitioning \emph{k}-space 
into label vs.\ mask sets. Table~\ref{tab:ssdu_configs} outlines the key hyperparameters, 
highlighting the uniform mask, $\ell_1$ \emph{k}-space loss, and typical Adam optimization 
(lr=$10^{-4}$).
\begin{table}[!ht]
\centering
\small
\caption{
Key hyperparameters used for training SSDU on knee and brain datasets.
}
\label{tab:ssdu_configs}
\begin{tabular}{lcc}
\toprule
\textbf{Hyperparameter}          & \textbf{Knee}          & \textbf{Brain} \\
\midrule
\multicolumn{3}{l}{\emph{Architecture}} \\[3pt]
Unrolled Steps                   & 4                      & 4                     \\
Block Architecture               & ResNet (256 ch)        & ResNet (256 ch)       \\
\# ResNet Blocks / Step          & 2                      & 2                     \\
Kernel Size                      & 3                      & 3                     \\
\midrule
\multicolumn{3}{l}{\emph{Self-supervision \& Masking Setup}} \\[3pt]
Mask Type                        & uniform                & uniform               \\
$\rho$ (Label Fraction)          & 0.2& 0.2\\
\midrule
\multicolumn{3}{l}{\emph{Training}} \\[3pt]
Loss Function                    & \emph{k}-space $\ell_1$ & \emph{k}-space $\ell_1$ \\
\textbf{Batch Size}             & 2                      & 1                     \\
\textbf{Grad. Accum. Steps}     & 8                      & 4                     \\
Base Learning Rate               & $10^{-4}$              & $10^{-4}$             \\
Weight Decay                     & $10^{-4}$              & $10^{-4}$             \\
Max Iterations                   & 80{,}000               & 80{,}000              \\
\bottomrule
\end{tabular}
\end{table}

\subsection{Reconstruction Results}
\label{appendix:recon_results}
All methods were trained and evaluated on \emph{clean}, non-noisy data. 
Table~\ref{table:recon_metrics_comparison_merged} lists the mean Peak Signal-to-Noise Ratio 
(PSNR) and Structural Similarity Index (SSIM) for knee and brain test sets across the six 
analyzed reconstruction methods. Overall, E2E-VarNet and MoDL achieve the highest PSNR values on knee 
scans, whereas N2R and VORTEX perform strongly on brain data—particularly evident in SSIM. 
U-Net and SSDU yield somewhat lower scores, potentially reflecting their more limited 
regularization strategies (fully data-driven in U-Net, fully self-supervised in SSDU). 
Notably, the reconstruction performances are comparable to those reported in the original 
studies or similar benchmarks, indicating that the networks were trained effectively. 
Overall, all methods surpass typical parallel imaging baselines in fidelity, confirming 
sufficient training convergence and general image quality for subsequent noise-variance 
analyses.

\begin{table}[t]
    \centering
    \caption{Quantitative reconstruction performance across methods on knee and brain datasets}
    \label{table:recon_metrics_comparison_merged}
    \begin{tabular}{lcccc}
\toprule
\textbf{Method} & \multicolumn{2}{c}{\textbf{Knee}} & \multicolumn{2}{c}{\textbf{Brain}} \\
\cmidrule(lr){2-3} \cmidrule(lr){4-5}
 & \textbf{PSNR (dB)} & \textbf{SSIM} & \textbf{PSNR (dB)} & \textbf{SSIM} \\
\midrule
E2E-VarNet & 40.52 & 0.899 & 36.31 & 0.895 \\
MoDL       & 40.60 & 0.899 & 36.44 & 0.847 \\
U-Net      & 38.81 & 0.852 & 36.12 & 0.849 \\
N2R        & 39.52 & 0.892 & 36.57 & 0.918 \\
VORTEX     & 40.47 & 0.894 & 35.78 & 0.914 \\
SSDU       & 39.88 & 0.862 & 35.51 & 0.885 \\
\bottomrule
\end{tabular}
\end{table}

\section{Computational Efficiency vs.\ Network Architecture}
\label{appendix:computation_vs_complexity}
Table~\ref{table:computation_times_scaled} compares \emph{empirical}  versus 
\emph{proposed} variance method computation times for different neural architectures on the knee dataset.  In unrolled frameworks (E2E-VarNet, N2R, VORTEX, SSDU), the computation 
time increases with more “Steps” (i.e., DC/regularization blocks), reflecting the deeper or more recurrent structure of the unrolled pipeline.  Nevertheless, our proposed method always remains an order of magnitude faster than empirical references. \emph{U-Net} requires fewer parameters (one-shot CNN without DC blocks), so the empirical approach is already 
relatively fast, yet we still observe a significant speedup with the proposed method.  Finally, \emph{MoDL} includes CG iterations within each DC step, driving up the empirical runtime significantly 
(181.5\,s for 4 steps), while our method’s runtime increases more modestly (36.9\,s).

These results indicate that both the \emph{unrolled depth} (number of repeated modules) and 
\emph{type of network layers} (e.g.\ CG-based DC, convolutional blocks, or potential self-attention 
units) can influence the total variance-computation time in our framework, albeit to a lesser 
extent than empirical MC.  For instance, substituting convolutional layers with 
self-attention could raise per-iteration overhead but might not drastically inflate the 
overall Jacobian-sketch cost, since our approach only requires \emph{Jacobian-vector products} 
rather than a full Jacobian matrix.  Ultimately, the proposed sketching method retains its 
computational advantage across diverse architectures, enabling efficient variance estimation 
without imposing strict constraints on layer choices or unrolling depth.

\begin{table}[t]
    \centering
    \small
    \caption{
    \textbf{Computation Times Across Different Neural Architectures.}
    Each entry lists the average time (in seconds) required to compute variance maps 
    per slice on the knee dataset, comparing \emph{Empirical} (MC) 
    versus our \emph{Proposed} (Jacobian-sketched) noise calculation. 
    Unrolled architectures vary in steps, indicating the number of DC/regularization blocks. 
    MoDL uses Conjugate Gradient (CG) steps within each DC block, and U-Net is a single-shot 
    CNN without explicit unrolled steps.
    }
    \label{table:computation_times_scaled}
    \setlength{\tabcolsep}{6pt}
    \renewcommand{\arraystretch}{1.2}
    \begin{tabular}{|l|c|c|c|}
        \hline
        \textbf{Architecture} & \textbf{Variant} & \textbf{Empirical time} & \textbf{Proposed time} \\
        \hline
        \multirow{5}{*}{Unrolled (E2E-VarNet, N2R, VORTEX, SSDU)} 
           & \textbf{2 Steps}  & 27.9  & 0.2 \\
           & \textbf{4 Steps}  & 54.0  & 1.3 \\
           & \textbf{6 Steps}  & 73.8  & 3.4 \\
           & \textbf{8 Steps}  & 97.8  & 12.1 \\
           & \textbf{10 Steps} & 120.6 & 20.6 \\
        \hline
        \textbf{U-Net}                    & \cite{unet}       & 9.6   & 0.4 \\
        \hline
        \textbf{Unrolled w/CG-DC (MoDL)}  & \textbf{4 Steps}  & 181.5 & 36.9 \\
        \hline
    \end{tabular}
\end{table}

\begin{figure*}[htb]
\vskip 0.2in
\begin{center}
\centerline{\includegraphics[width=\columnwidth]{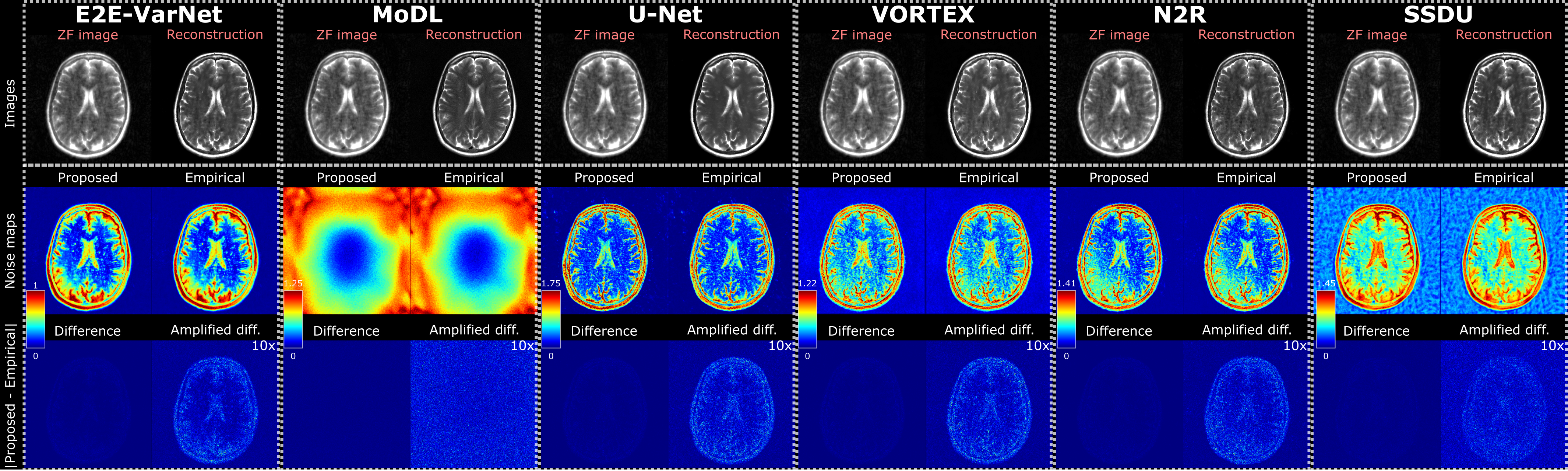}}
\caption{Each column corresponds to a distinct deep reconstruction 
    method (E2E-VarNet, MoDL, U-Net, VORTEX, N2R, SSDU) on brain data at $R=8$, $\alpha=1$. In each column:
    \textbf{(top row)} shows the zero-filled (ZF) image and the final reconstructed;
    \textbf{(middle row)} displays variance maps derived by the proposed
    method and empirical simulations;
    \textbf{(bottom row)} presents the difference map between the proposed and empirical variance 
    estimates, plus an “amplified” difference to further highlight spatial discrepancies. The color bars indicate each map’s display window in arbitrary units of variance. Because MoDL integrates an iterative congute-gradient CG-based data consistency step that repeatedly applies the imaging and adjoint operator, the resulting noise map often mirrors the specific coil geometry and undersampling scheme. Regions with poorer coil coverage or unsampled k‐space lines accumulate higher residuals, as the iterative solver relies more heavily on the learned prior in those areas. Consequently, MoDL’s noise distribution visibly reflects the interplay of coil sensitivities, the mask pattern, and the iterative gradient updates compared to data-dependent Jacobian structure, resulting in the distinct patterns
    }
\label{fig:brain_methods}
\end{center}
\vskip -0.2in
\end{figure*}

% \section{Results for Effect of Undersampling}
\begin{figure*}[htb]
\vskip 0.2in
\begin{center}
\centerline{\includegraphics[width=\columnwidth]{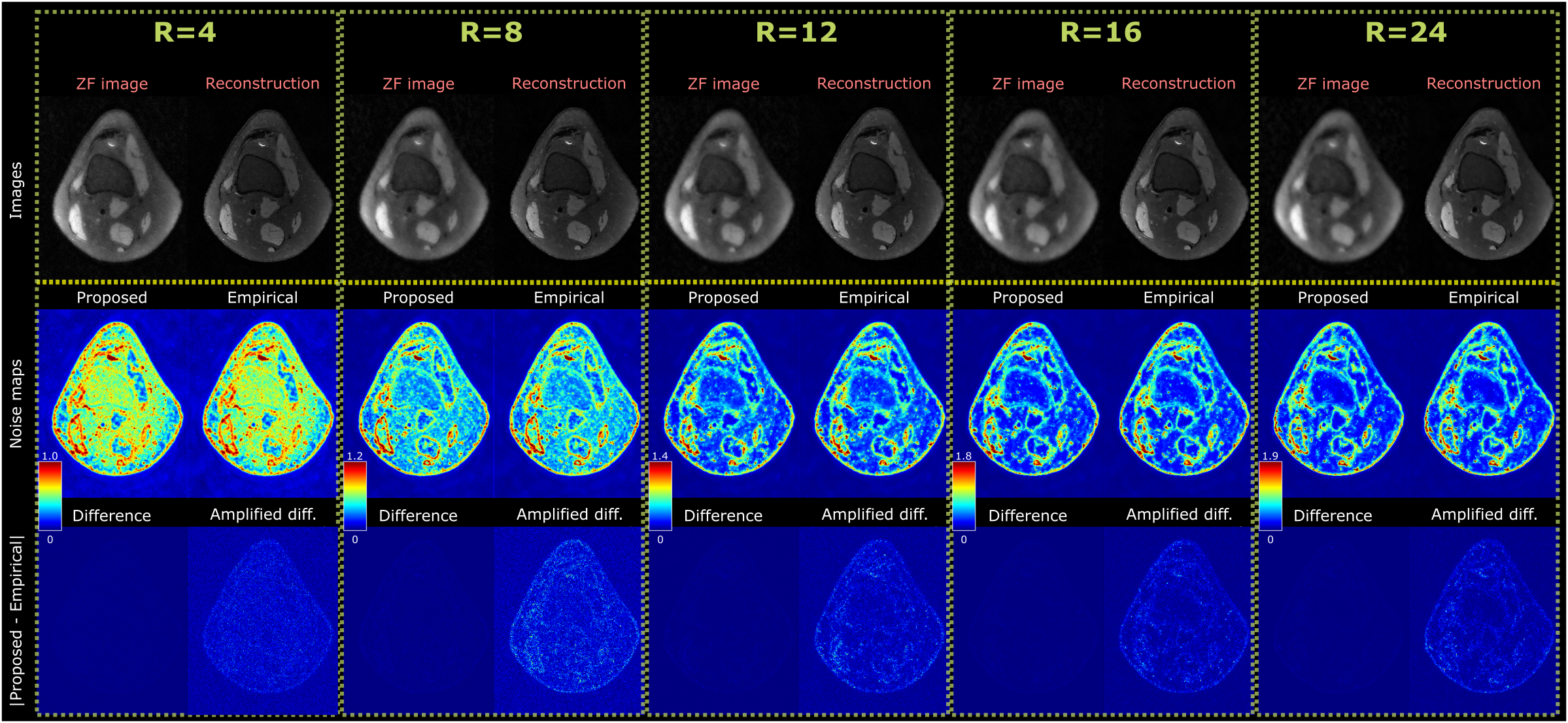}}
\caption{Each column corresponds to a different acceleration factor $R$ (MoDL on knee at $\alpha=1$). In each column:
    \textbf{(top row)} shows the zero-filled (ZF) image and the final reconstructed;
    \textbf{(middle row)} displays variance maps derived by the proposed
    method and empirical simulations;
    \textbf{(bottom row)} presents the difference map between the proposed and empirical variance 
    estimates, plus an “amplified” difference to further highlight spatial discrepancies. The color bars indicate each map’s display window in arbitrary units of variance. 
    }
\label{fig:knee_R_variance_maps}
\end{center}
\vskip -0.2in
\end{figure*}

\begin{figure*}[ht]
\vskip 0.2in
\begin{center}
\centerline{\includegraphics[width=\columnwidth]{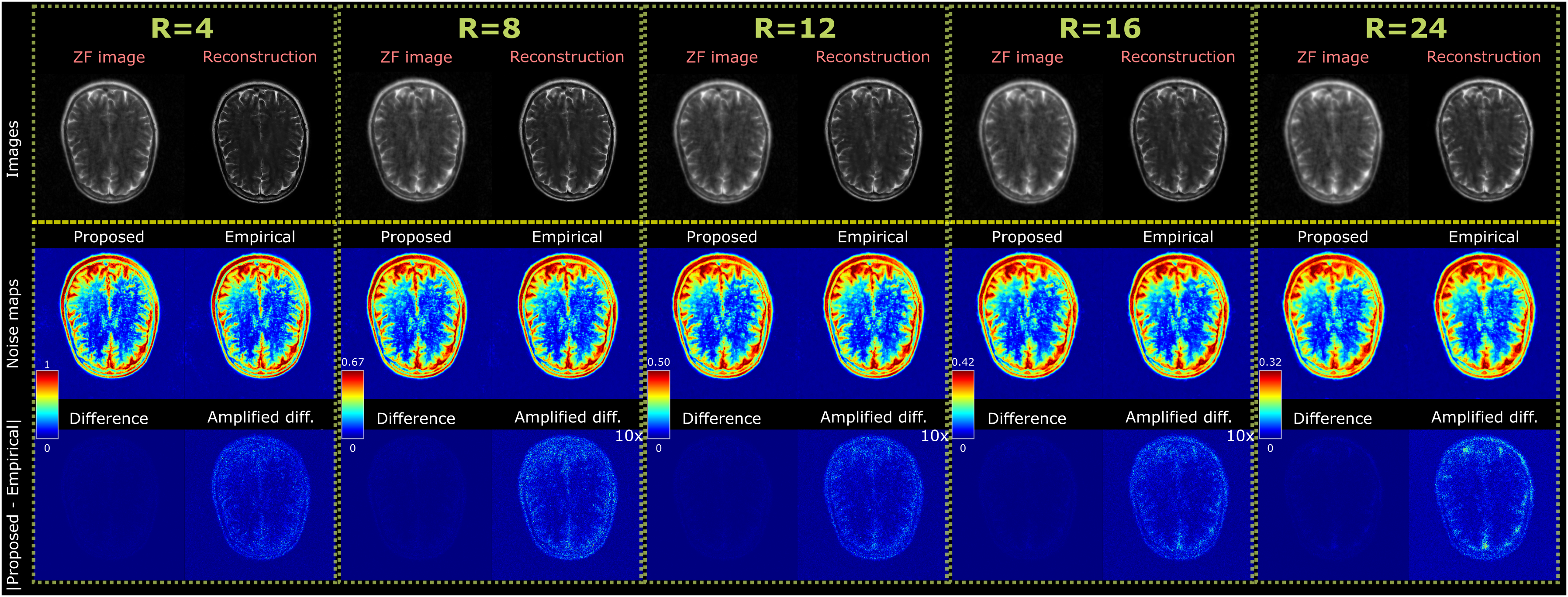}}
\caption{Each column corresponds to a different acceleration factor $R$ (E2E-VarNet on brain at $\alpha=1$). In each column:
    \textbf{(top row)} shows the zero-filled (ZF) image and the final reconstructed;
    \textbf{(middle row)} displays variance maps derived by the proposed
    method and empirical simulations;
    \textbf{(bottom row)} presents the difference map between the proposed and empirical variance 
    estimates, plus an “amplified” difference to further highlight spatial discrepancies. The color bars indicate each map’s display window in arbitrary units of variance.}
\label{fig:brain_R_variance_maps}
\end{center}
\vskip -0.2in
\end{figure*}

\section{Practical Comparison with Low-Trial Monte Carlo}
\label{appendix:mc_vs_sketching}
\begin{figure}[htb]
    \centering
    \includegraphics[width=\linewidth]{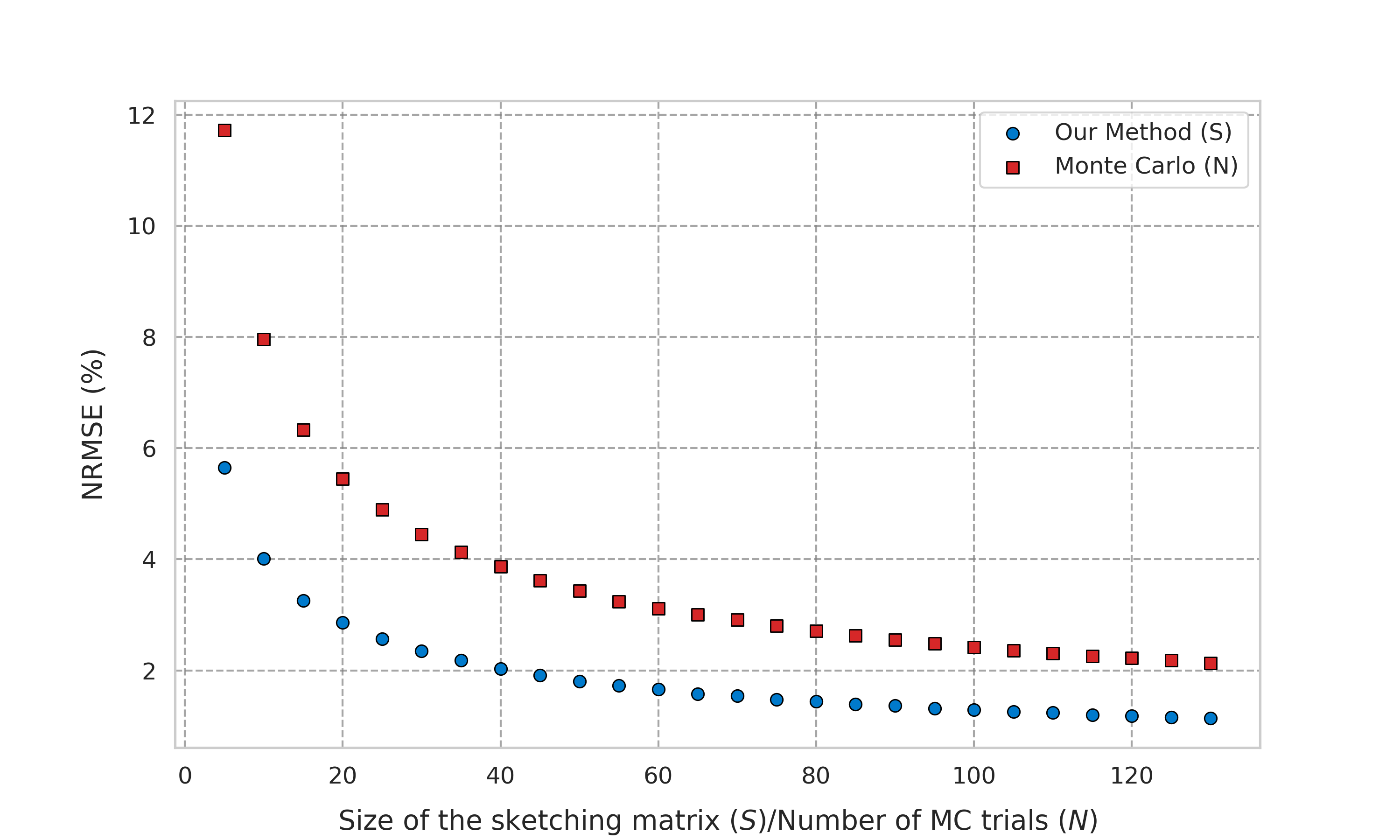}
    \caption{
    Comparison of the proposed method with \(S\) (sketching matrix size) and empirical baseline with \(N\) (MC trials).
    NRMSE (\%) is measured against a high-accuracy reference map obtained with \(N=10{,}000\) (VORTEX on brain at $\alpha=1$). 
    For small sample sizes (\(\leq 50\)), Monte Carlo exhibits markedly higher error, while our method yields lower error at all examined sketch sizes. Even at \(S=N=100\), the proposed approach offers less error with an order-of-magnitude reduction in computational time (\(0.12\)s vs.\ \(1.8\)s for MC). 
    }
    \label{fig:sketching_vs_mc}
\end{figure}

A natural question arises regarding whether empirical reference maps with smaller number of MC trials (e.g., 50--100) might suffice to achieve competitive accuracy without incurring the high computational cost typically associated with MC sampling. To explore this, we compared the proposed sketching method (with sketch size \(S\)) and MC (with \(N\) trials), each varied over a range from 5 to 130, and measured the NRMSE against a high-accuracy reference map computed using \(N=10{,}000\). Experiments were performed on E2E-VarNet at $R=8,\alpha=1$. As shown in Figure~\ref{fig:sketching_vs_mc}, MC exhibits large estimation errors for \(N\lesssim 50\), whereas our method maintains relatively lower errors even at small \(S\). At larger values of \(S\approx N\), our approach remains more accurate while also being faster to compute: for instance, at \(S=N=100\), MC requires \(\sim\!1.8\) seconds, whereas our method takes only \(\sim\!0.12\) seconds. This behavior aligns with our convergence experiments (see \ref{appendix:monte}), indicating that MC variance estimates tend to require three orders of magnitude trials (e.g., \(N\approx1{,}000\)–\(10{,}000\)) to reach a reliable accuracy. In contrast, the sketching-based method attains lower error with at very few sample regimes. Hence, even when MC is reduced to 50--100 trials to speed up computation, the resulting noise estimates suffer in accuracy compared to the proposed sketching approach, highlighting a clear advantage in both performance and speed.

\section{Undersampling Patterns}
\label{sec:undersampling_patterns}
\begin{figure*}[ht]
\vskip 0.2in
\begin{center}
\centerline{\includegraphics[width=\columnwidth]{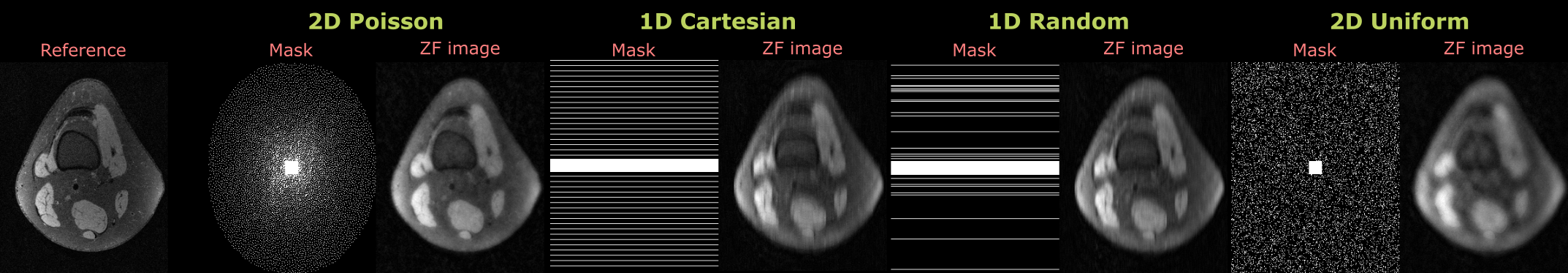}}
\caption{Different undersampling schemes analyzed in this study.  For a representative slice, ZF images generated by the undersampling masks are showcased, along with the fully-sampled reference. Coherent and incoherent
aliasing artifacts are generally exhibited by 1D and 2D undersampling masks, respectively.}
\label{fig:undersampling_masks}
\end{center}
\vskip -0.2in
\end{figure*}

In this study, we analyzed a range of undersampling schemes to evaluate their impact on noise propagation and variance estimation. These include sampling patterns widely used in musculoskeletal imaging, such as 1D Cartesian uniform and random undersampling, as well as  2D pseudo-random schemes like Poisson Disc and uniform random sampling \cite{cartesian, skm_tea} (See Fig. \ref{fig:undersampling_masks}. Each pattern introduces distinct aliasing artifacts in the image domain—ranging from coherent aliasing artifacts in uniform Cartesian sampling to incoherent optimized for parallel imaging.
to noise-like incoherent artifacts in 2D pseudo-random patterns that are incoherent patterns are better suited for DL algorithms, while Cartesian sampling is often optimized for parallel imaging. Here we investigate these diverse schemes to provide insights into how different undersampling strategies influence the efficacy our noise calculation method (\ref{subsec:undersampling_effect}).

\begin{figure*}[ht]
% \vskip 0.2in
\begin{center}
\centerline{\includegraphics[width=\columnwidth]{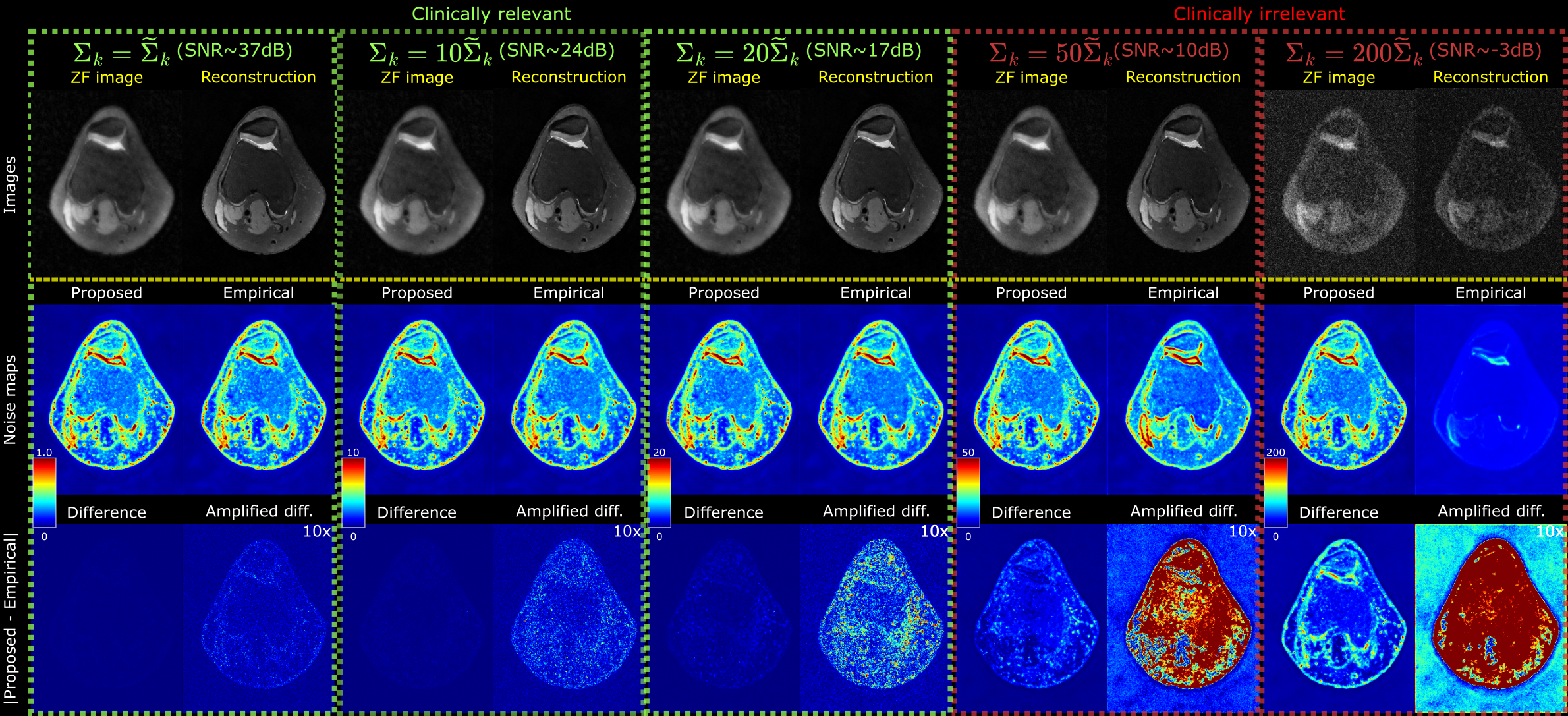}}
\caption{Each column corresponds to a different noise scaling factor $\alpha$ resulting in varying SNR scenarios (E2E-VarNet on knee, at $R=8$). In each column:
    \textbf{(top row)} shows the zero-filled (ZF) image and the final reconstructed;
    \textbf{(middle row)} displays variance maps derived by the proposed
    method and empirical simulations;
    \textbf{(bottom row)} presents the difference map between the proposed and empirical variance 
    estimates, plus an “amplified” difference to further highlight spatial discrepancies. The color bars indicate each map’s display window in arbitrary units of variance. Note that $\alpha=50$–$200$ corresponds to SNR values below $\approx10\,\mathrm{dB}$ 
    (i.e., well under an SNR of 10-15 that is often cited as a threshold for diagnostic utility.}
\label{fig:noise_analysis_knee}
\end{center}
\vskip -0.4in
\end{figure*}

\section{Additional Results for Robustness to k-space Noise Level}
\label{appendix:noise_discussion}
\begin{figure}[ht]
\vskip 0.2in
\begin{center}
\centerline{\includegraphics[width=\columnwidth]{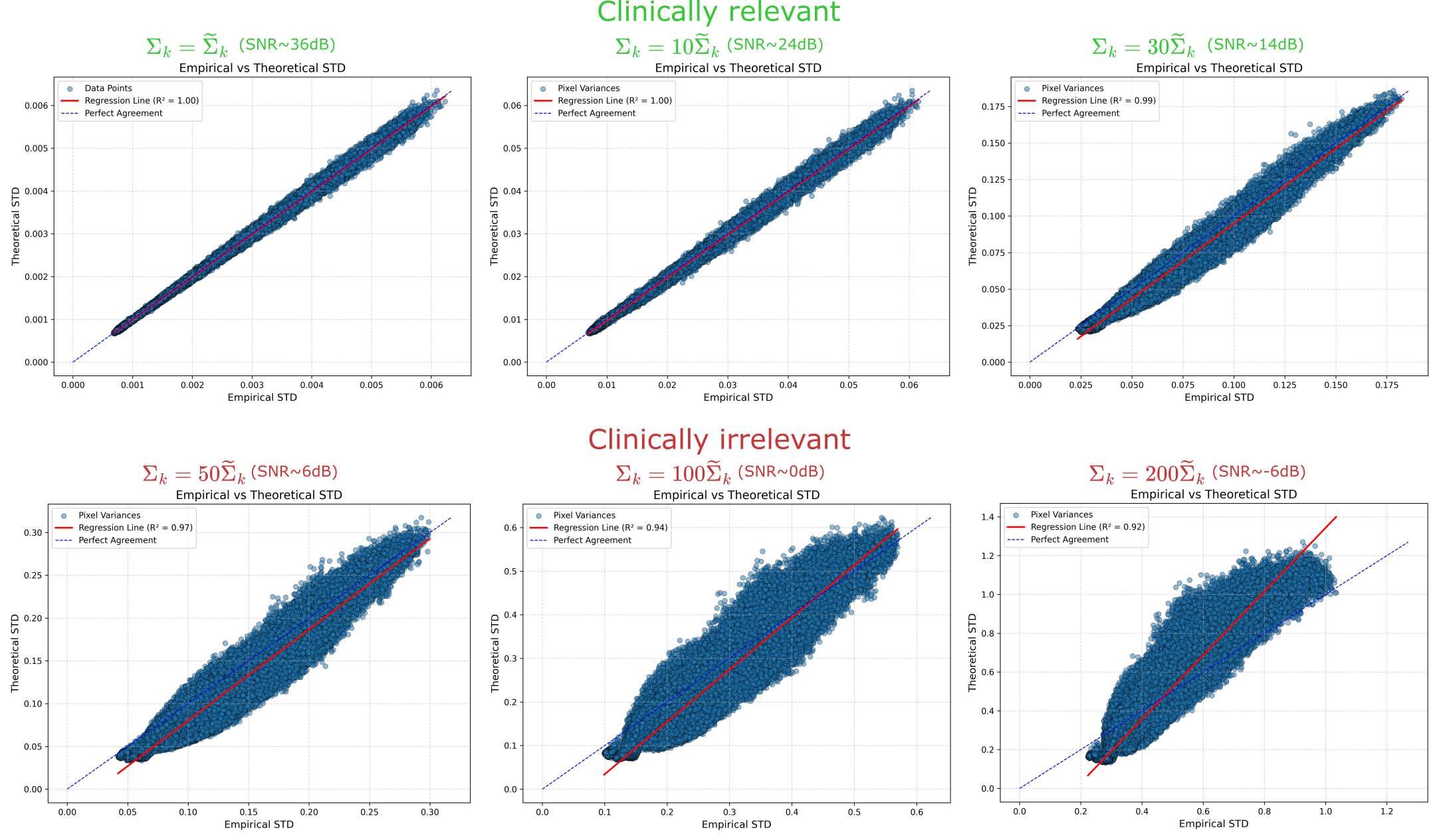}}
\caption{For a representative brain slice, empirical
variances (horizontal axes) were compared to variances calculated our method (vertical 
axes) under successively larger noise scales: \( \alpha= 1\, 10, 30, 50, 
100,200\) (E2E-VarNet, at $R=8$). In each panel, the diagonal line represents 
ideal one-to-one correspondence between empirical and calculated estimates, while the solid 
line indicates the best-fit linear regression (with its $R^2$ value indicated). Each data point 
 corresponds to a single voxel’s empirical vs.\ calculated variance, visualizing how closely they match at different noise levels.}
\label{fig:empirical_vs_theoretical_std}
\end{center}
\vskip -0.2in
\end{figure}

Figure~\ref{fig:empirical_vs_theoretical_std} demonstrates tight alignment around 
the diagonal, exhibiting near‐unity \(R^2\) values that confirm a strong match between empirical 
and theoretical STD at \emph{lower to high noise} (\(1\times\)--\(30\times\)). As noise gets excessively larger to the clinically irrelevant regimes (\(50\times\)--\(200\times\)), our model \emph{overestimates} the output noise
relative to the reference: the best‐fit regression line lies above the diagonal, and the 
scatter broadens. Nevertheless, the correlation remains consistently high, 
indicating that the model still captures the main trend in noise propagation while modestly 
overestimating variance in these extreme scenarios.

\begin{figure}[t!]
\vskip 0.2in
\begin{center}
\centerline{\includegraphics[width=\columnwidth]{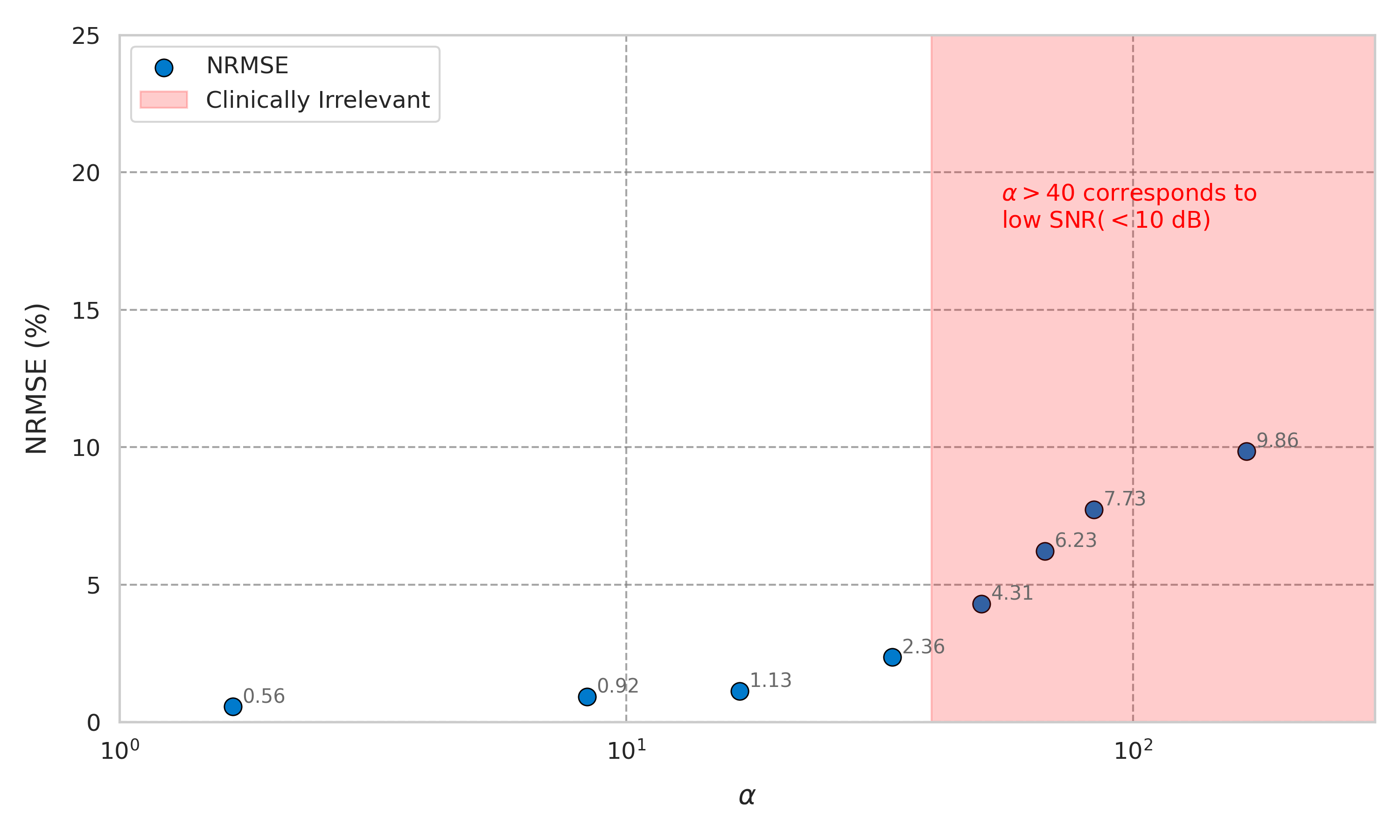}}
\caption{Mean NRMSE between reference empirical variance maps and variance maps from our method across test slices were plotted against increased noise levels (E2E-VarNet on knee, at $R=8$). Here, $\alpha$ is plotted on a log scale and progressively scales the noise covariance matrix 
    $\bm{\Sigma}_k = \alpha \widetilde{\bm{\Sigma}}_k$, simulating higher noise, and accordingly lower SNR 
    levels.}
\label{fig:nrmse_vs_noise}
\end{center}
\vskip -0.2in
\end{figure}

%%%%%%%%%%%%%%%%%%%%%%%%%%%%%%%%%%%%%%%%%%%%%%%%%%%%%%%%%%%%%%%%%%%%%%%%%%%%%%%
%%%%%%%%%%%%%%%%%%%%%%%%%%%%%%%%%%%%%%%%%%%%%%%%%%%%%%%%%%%%%%%%%%%%%%%%%%%%%%%

\end{document}